\pdfoutput=1
\documentclass[11pt]{article}
\usepackage[final]{acl}

\usepackage{times}
\usepackage{latexsym}

\usepackage{xspace}
\usepackage[T1]{fontenc}
\usepackage[utf8]{inputenc}

\usepackage{microtype}

\usepackage{inconsolata}

\usepackage{pifont}
\usepackage{times}
\usepackage{latexsym}
\usepackage{booktabs}
\usepackage{algorithm}
\usepackage{algorithmic}
\usepackage{multicol}
\usepackage{amsfonts}
\usepackage{graphicx}
\usepackage{booktabs}
\usepackage{multirow}
\usepackage{xcolor}
\usepackage{makecell} % 表格单元格内容换行
\usepackage{array} % 修改表格行间距
\usepackage{fontawesome} % 插入 FontAwesome 图标（如 \faToggleOff 和 \faUser）
\usepackage{xcolor} % 可选：表格颜色自定义
\usepackage{caption} % 控制标题格式
\usepackage{amssymb} % 可选：扩展数学符号
\usepackage{amsmath} % 可选：数学公式支持
\usepackage{colortbl}
\usepackage{ulem} % 用于处理下划线
\usepackage[listings,skins,breakable]{tcolorbox}
\usepackage{pifont}

\usepackage{float} % 如果使用 [H] 选项

\definecolor{dkgreen}{rgb}{0,0.6,0}
\lstdefinelanguage{prompt}{
    basicstyle=\scriptsize\ttfamily,
    keywordstyle=\color{blue},
    keywords={class, dspy},
    moredelim = [s][\color{dkgreen}]{"""}{"""},
}
\newtcblisting[use counter=promptcounter]{prompt}[1][]{
  enhanced,
  arc=0em,
  boxrule=.5pt,
  listing only,
  breakable,
  listing options={
    language=prompt,
    upquote=true,
    basicstyle=\scriptsize \ttfamily \setlength{\baselineskip}{1.1\baselineskip},
    breaklines=true,
    breakindent=8pt,
    xleftmargin=-8pt,
    xrightmargin=-8pt,
    aboveskip=-4pt,
    belowskip=-4pt,
    columns=fullflexible,
    escapeinside={|}{|}
  },
  colback=white,
  colframe=gray,
  colbacktitle=gray!5,        % 设置标题背景色
  coltitle=black,             % 设置标题文字颜色
  attach boxed title to top center={yshift=-3mm},
  #1
}

\lstdefinestyle{mystyle}{
    backgroundcolor=\color{backcolour},   
    commentstyle=\color{codegreen},
    keywordstyle=\color{magenta},
    numberstyle=\tiny\color{codegray},
    stringstyle=\color{codepurple},
    basicstyle=\ttfamily\footnotesize,
    breakatwhitespace=false,         
    breaklines=true,                 
    captionpos=b,                    
    keepspaces=true,                 
    numbers=left,                    
    numbersep=5pt,                  
    showspaces=false,                
    showstringspaces=false,
    showtabs=false,                  
    tabsize=2
}

\definecolor{aliceblue}{RGB}{255, 238, 241}

\definecolor{babyred}{rgb}{0.85, 0.93, 0.97}

\definecolor{uclablue}{RGB}{159, 195, 224}

\definecolor{uclagold}{RGB}{255, 240, 180}

\definecolor{grayred}{RGB}{205, 186, 207}

\definecolor{mygray}{gray}{.90}

\newcommand{\github}{\raisebox{-1.5pt}{\includegraphics[height=1.05em]{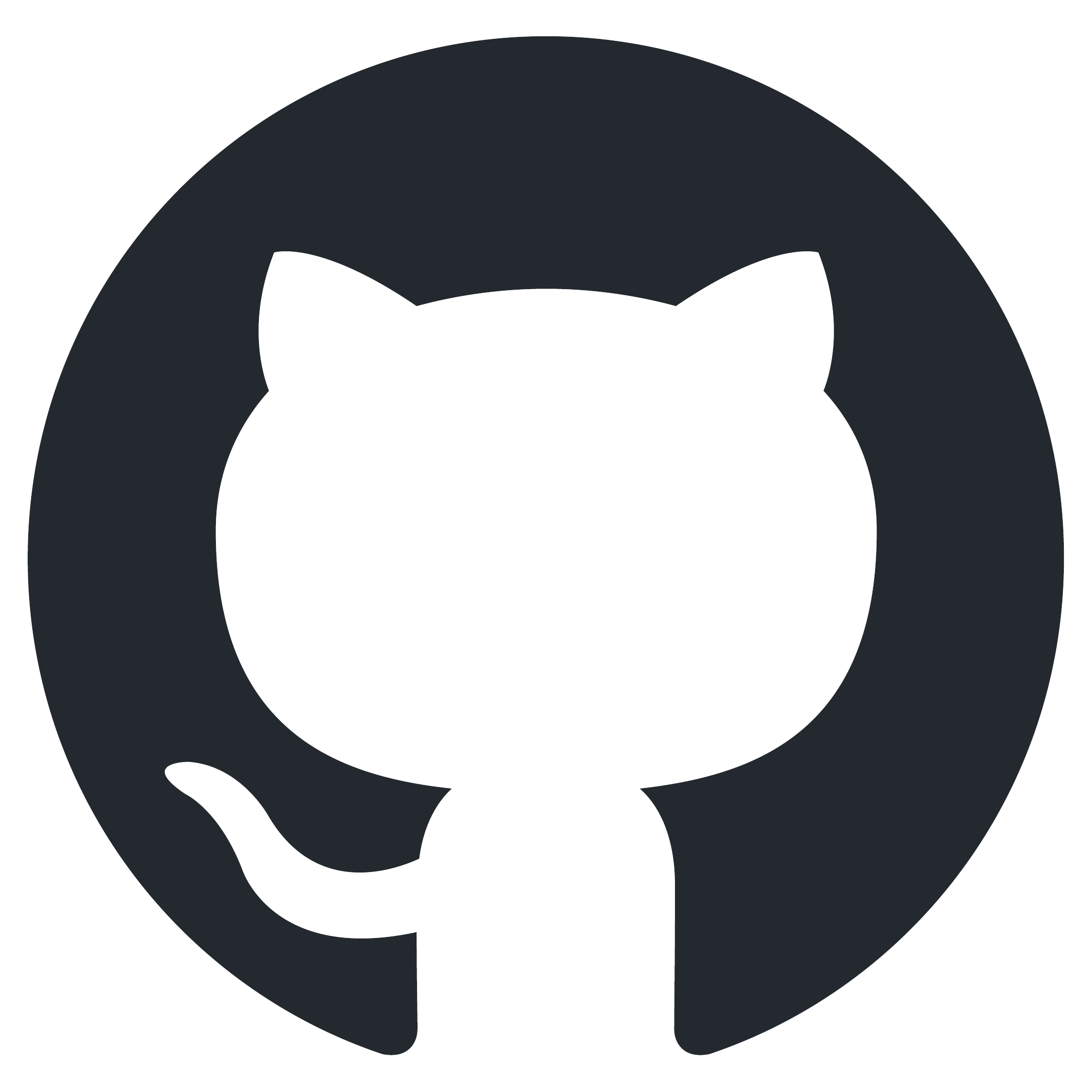}}\xspace}

\newcommand{\modelscope}{\raisebox{0pt}{\includegraphics[height=0.8em]{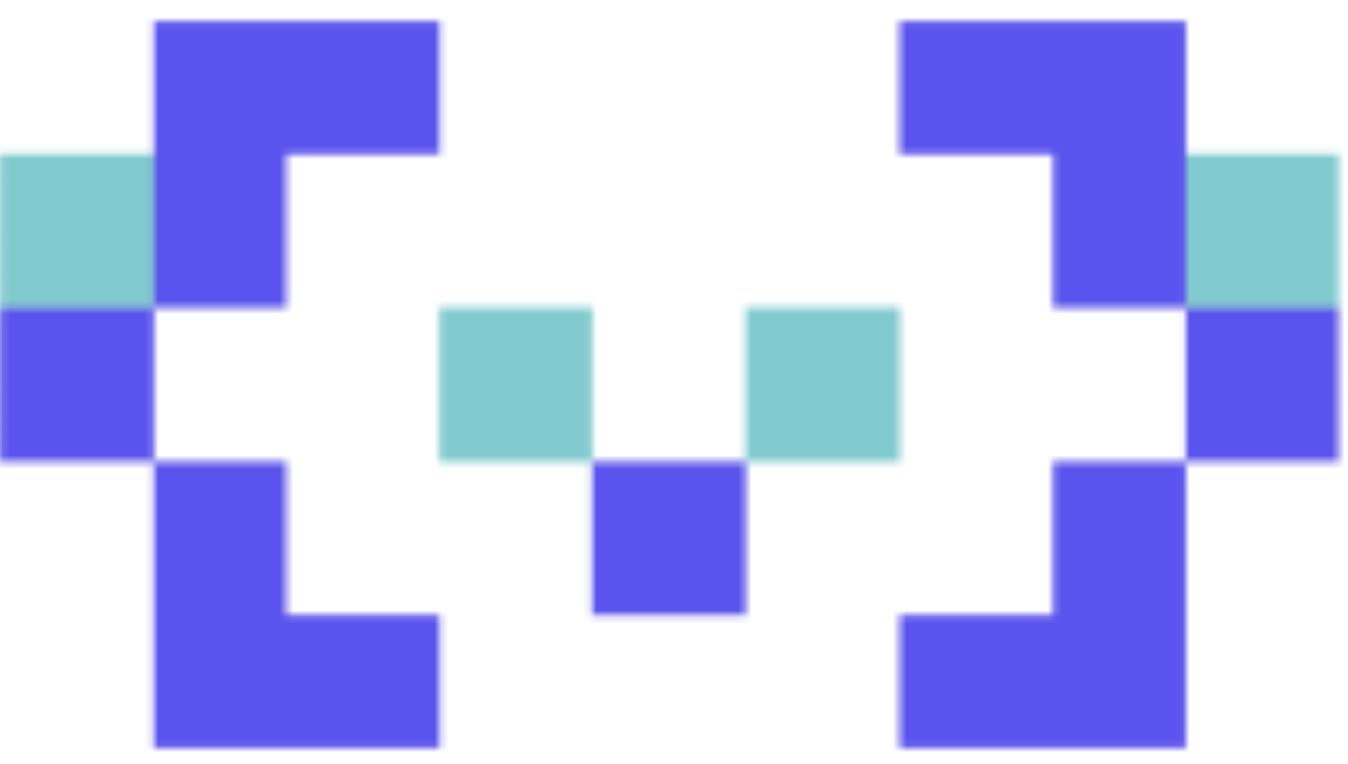}}}

\newcommand{\link}{\raisebox{0pt}{\includegraphics[height=0.9em]{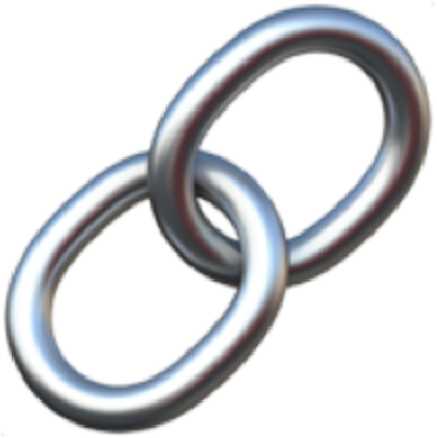}}\xspace}

\author{
    \textbf{Zekun Xi$^{{1,2}}$},\hspace{0.5mm}
    \textbf{Wenbiao Yin$^2$},\hspace{0.5mm}
    \textbf{Jizhan Fang$^1$},\hspace{0.5mm}
    \textbf{Jialong Wu$^2$},\hspace{0.5mm}\hspace{0.5mm}
    \textbf{Runnan Fang$^{{1,2}}$},\\
    \textbf{Yong Jiang$^{2*}$},\hspace{0.5mm}
    \textbf{Pengjun Xie$^2$},\hspace{0.5mm}
    \textbf{Fei Huang$^2$},\hspace{0.5mm}
    \textbf{Huajun Chen$^{1,3}$},\hspace{0.5mm}
    \textbf{Ningyu Zhang$^{1,3}$}\thanks{~~Corresponding Author.}\hspace{0.5mm}\\
    $^1$Zhejiang University\\
    $^2$Tongyi Lab, Alibaba Group\\
    $^3$Zhejiang Key Laboratory of Big Data Intelligent Computing\\
    \texttt{\{xizekun2023, zhangningyu\}@zju.edu.cn} \\
    \vspace{.5em}\link \href{https://zjunlp.github.io/project/OmniThink}{Homepage}
    \vspace{.5em}\modelscope \href{https://www.modelscope.cn/studios/iic/OmniThink}{Demo} 
    \vspace{.5em}\github \href{https://github.com/zjunlp/OmniThink}{Code}
} 

\begin{document}
\title{OmniThink: Expanding Knowledge Boundaries in Machine Writing through Thinking}
 
% 后续补上outline对于文章质量的评估
% 第二章节画一个表讲清楚和co-storm的差别
% 补上conceptual pool和information tree的消融
% 再补上一个knowledge density的人类偏好对齐的指标
% 最后补上文章长度的分析，还有token开销的
% 实验方差也补上
% 把什么时候停止延展也讲清楚

\maketitle
\begin{abstract}
Machine writing with large language models often relies on retrieval-augmented generation. However, these approaches remain confined within the boundaries of the model's predefined scope, limiting the generation of content with rich information. Specifically, vanilla-retrieved information tends to lack depth, novelty, and suffers from redundancy, which negatively impacts the quality of generated articles, leading to shallow, unoriginal, and repetitive outputs. To address these issues, we propose OmniThink, a slow-thinking machine writing framework that emulates the human-like process of iterative expansion and reflection. The core idea behind OmniThink is to simulate the cognitive behavior of learners as they slowly deepen their knowledge of the topics. Experimental results demonstrate that OmniThink improves the knowledge density of generated articles without compromising metrics such as coherence and depth. Human evaluations and expert feedback further highlight the potential of OmniThink to address real-world challenges in the generation of long-form articles.
\end{abstract}

\section{Introduction}

% \begin{quote}
% ``\textit{Education is not the learning of facts, but the training of the mind to think.}''\\
% \hspace*{\fill}--- Albert Einstein
% \end{quote}

% \begin{quote}
% ``\textit{We can only see a short distance ahead, but we can see plenty there that needs to be done.}''\\
% \hspace*{\fill}--- Alan Turing
% \end{quote}

\begin{figure}[h]
    \centering
    \includegraphics[width=0.95\linewidth]{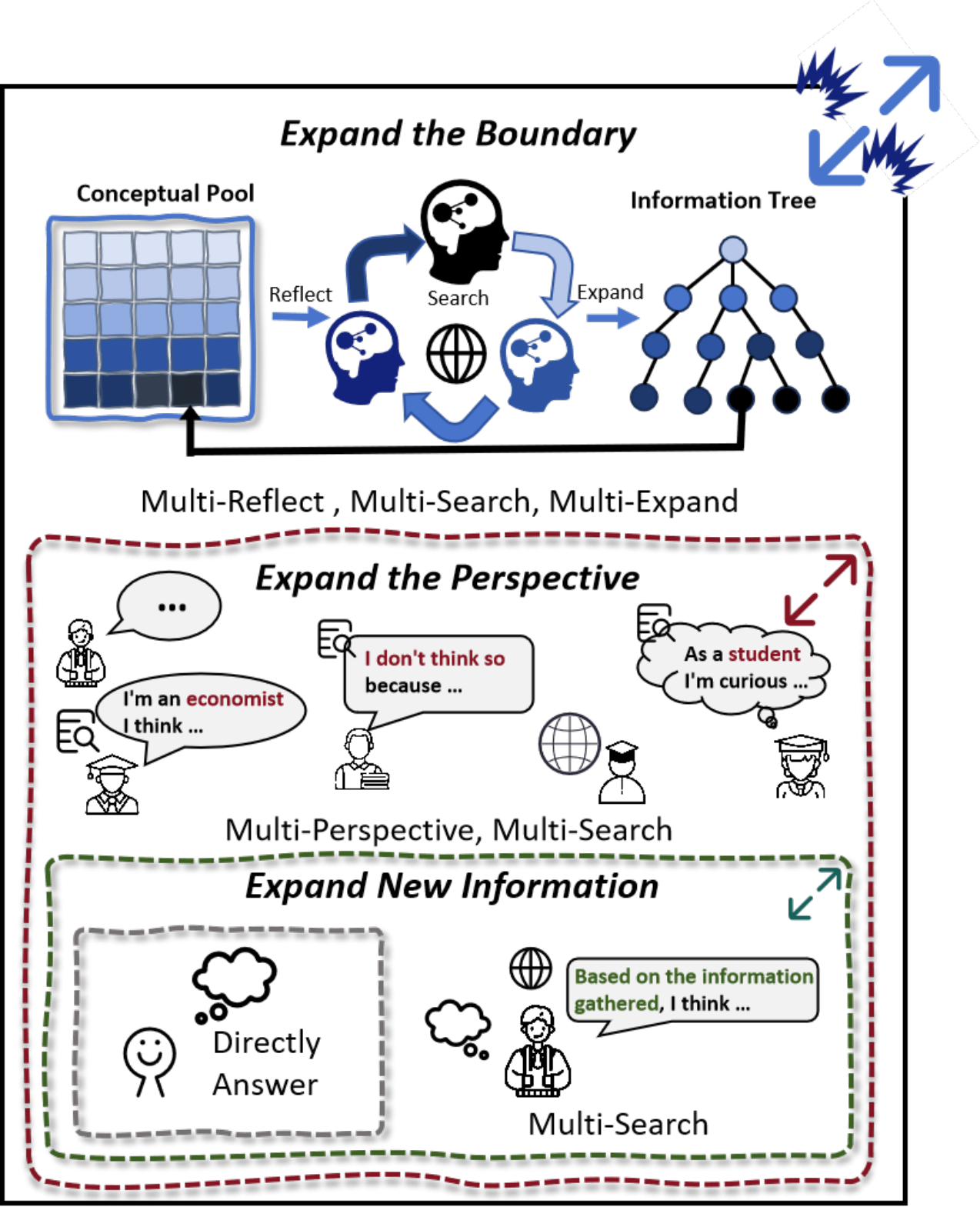}
    \caption{
    Previous machine writing approaches only expand new information or perspective via RAG and role-playing. 
    OmniThink expands knowledge boundaries through continuous reflection and exploration, attaching knowledge to an information tree and extracting it into a conceptual pool to deepen understanding and uncover more in-depth content.
    }
    \label{fig:example}
\end{figure}

Writing is a continuous process of collecting information and thinking \cite{bean2021engaging}.
Recent advances in Large Language Models (LLMs) have demonstrated remarkable progress in machine writing such as open domain long-form generation~\cite{liang-etal-2023-open, yang2023docimprovinglongstory,zhao2024surveylargelanguagemodels} or report generation on specific topics~\cite{liu2018generatingwikipediasummarizinglong}. 
To seek useful information, as shown in Figure \ref{fig:example}, early attempts use Retrieval Augmented Generation (RAG) to \emph{expand new information} on a given topic \cite{gao2024retrievalaugmentedgenerationlargelanguage,edge2024local}.
However, vanilla RAG relies on a fixed set of search strategies~\cite{ram2023incontextretrievalaugmentedlanguagemodels}, which lack diversity in generation, preventing a thorough exploration of the topic and resulting in a fragmented and incomplete understanding of the subject~\cite{spink1998highly}.
To address this issue, STORM~\cite{shao2024assisting} and Co-STORM~\cite{jiang2024unknownunknowns} have proposed a role-play approach designed to \emph{expand the perspective}, which means collecting information from multiple perspectives, thus broadening the information space~\cite{shen2023summarizationdesigningaisupport,shanahan2023role,parmar2010stakeholder}.
Yet these approaches are still being thought within the scope of one's own role, making it difficult to generate deep content and break through one's own knowledge boundaries\cite{ji2025testtimecomputingsystem1thinking}.
In particular, retrieved information often lacks depth, novelty and redundancy, directly affecting the quality of generated articles, resulting in shallow, repetitive, and unoriginal outputs~\cite{skarlinski2024languageagentsachievesuperhuman}.

Note that humans can naturally avoid such pitfalls in the writing process. 
This phenomenon can be explained through the theory of reflective practice, a concept rooted in cognitive science~\cite{osterman1990reflective}. 
According to this theory, human writers continuously reflect on previously gathered information and personal experiences, allowing them to reorganize, filter, and refine their cognitive framework. 
This process prompts writers to iteratively adjust their writing direction and mental pathways, ultimately allowing human authors to generate more profound, nuanced and original content \cite{bruce1978cognitive}.

Motivated by this, we propose OmniThink, a new machine writing framework that emulates the human-like cognitive process. 
The core idea behind OmniThink is to simulate the cognitive behavior of learners as they gradually deepen their understanding of complex topics to \emph{expand knowledge boundaries}.
We introduce two innovative components, information tree and conceptual pool, to simulate the process of collecting information and structuring cognition during human iterative learning. 
Through continuous expansion and reflection, these components are enriched.
% By continuously reflecting on previously retrieved information, OmniThink can determine the optimal steps for further expansion. 
% This expansion-reflection mechanism enables the dynamic adjustment of the retrieval strategies, fostering a more thorough and comprehensive exploration of relevant information.
Once a diverse set of information has been gathered and structured, OmniThink transitions to the stages of outline construction and article generation. 
This iterative thinking process leads to the production of articles of higher quality that contain a higher knowledge density of useful, insightful, and original content.
OmniThink is model-agnostic and can be integrated with existing frameworks. 

We evaluate OmniThink on the WildSeek datasets~\cite{jiang2024unknownunknowns} based on previous metrics as well as a new metric, named knowledge density.
Experimental results demonstrate that OmniThink enhances the knowledge density of generated articles without compromising key metrics such as coherence and depth. 
% Human evaluations and expert feedback further underscore the potential of our approach in addressing real-world challenges in the the generation of long-form articles.
To conclude, our main contributions are as follows:

\begin{itemize}
    \item We propose OmniThink, a novel writing framework that emulates the human slow-thinking process. 
    \item We propose a new metric, Knowledge Density (KD), which measures the proportion of useful information in an article.
    \item We analyze the challenges of current long-form generation methods from a novel knowledge boundary perspective, investigate the underlying factors contributing to the effectiveness of OmniThink, and propose a new direction for future long-form generation research.
\end{itemize}

\section{Background}
\label{sec:bg}

\subsection{Task Definition}
\label{sec:definition}
We focus on the task of open-domain long-form generation for machine writing, which retrieving information from an open domain and synthesizing it into a coherent article ~\cite{fan2019eli5,su2022readgeneratefaithfullong,quan2024languagemodelsselflengthengenerate}. 
Given an input topic $\mathrm{T}$, the target of open-domain long-form generation is to generate a long article $\mathcal{A}$. 
The current standard approach involves two major steps~\cite{zhang2019outline,zheng2023outline}:
\textit{(i)} Use a search engine $\mathcal{S}$ to retrieve information $\mathcal{I}=\mathcal{S}(\mathrm{T})$ which is related to the topic $\mathrm{T}$; 
\textit{(ii)} Generate an outline $O = \text{Generate}(\mathcal{I}, \mathrm{T})$ based on the retrieved information $\mathcal{I}$ and input topic $\mathrm{T}$.
Finally, the article is generated using the outline, expressed as $\mathcal{A} = \text{Generate}(O, \mathcal{I})$.

\subsection{Revisiting Previous Methods}
\begin{figure}[ht]
    \centering
    \includegraphics[width=0.95\linewidth]{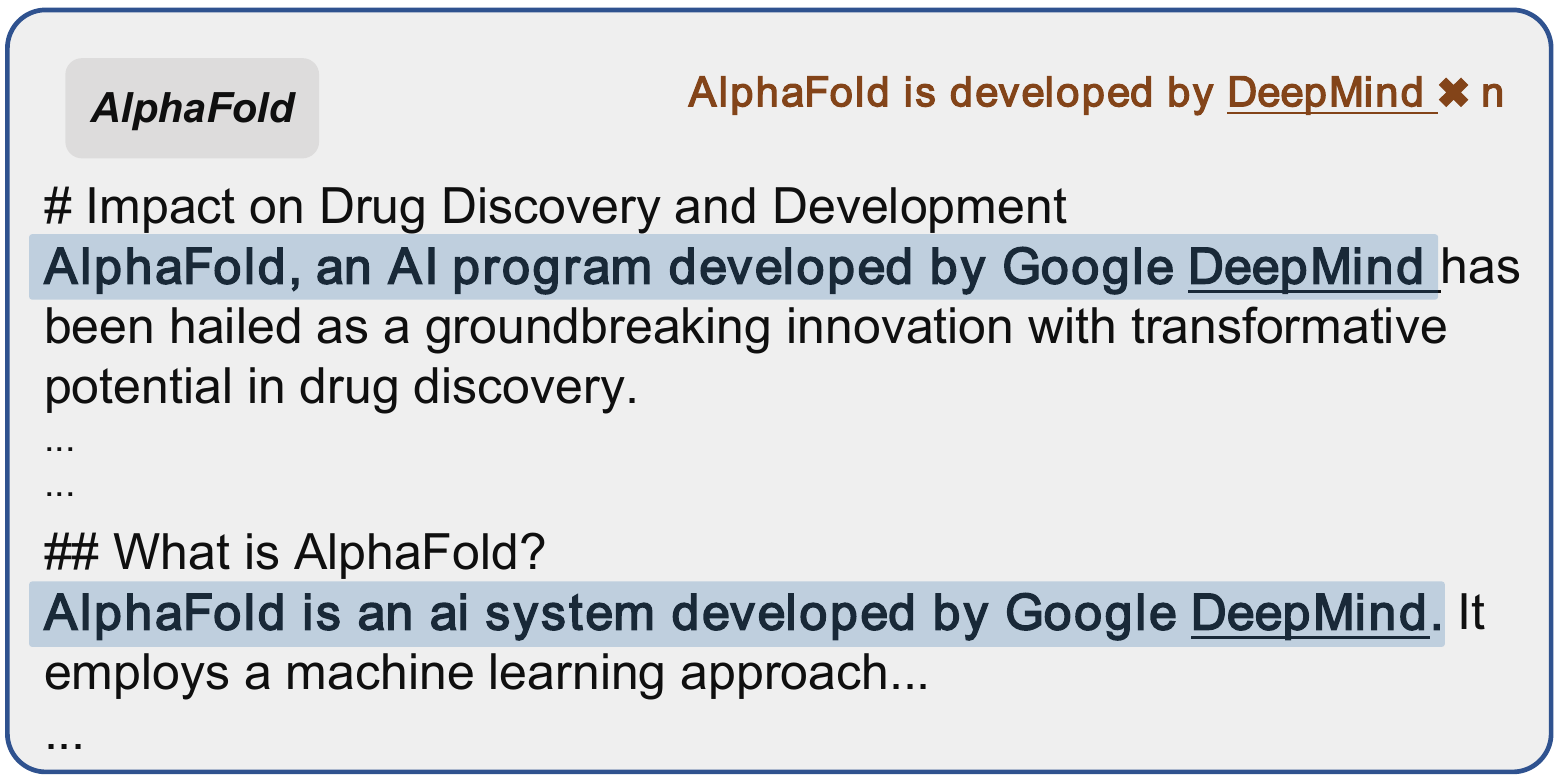}
    \caption{
    A case generated by STORM using GPT-4o on the topic of AlphaFold.    
    We have marked the repeated expressions in the article regarding ``AlphaFold is developed by DeepMind''.
    }
    \label{fig:case}
\end{figure}
Previous works have made numerous efforts to improve the quality of open-domain long-form generation.
Co-STORM~\cite{jiang2024unknownunknowns} introduces a user-participatory roundtable discussion in step \textit{(i)} to enhance the diversity of the retrieved information.
STORM~\cite{shao2024assisting} proposes a questioning mechanism to improve the quality and relevance of the generated outlines in step \textit{(ii)}.

Although substantial progress has been made in open-domain long-form generation, a persistent challenge remains: the generated content frequently suffers from \textbf{redundancy} and \textbf{lacks novelty}.
We present a case generated by STORM~\cite{shao2024assisting} with GPT-4o as the backbone, as shown in Figure \ref{fig:case}. 
In this article, the well-known phrase ``AlphaFold was developed by DeepMind'' appears multiple times, whereas it could be stated only once in the initial mention.

\subsection{Limitation Analysis From A Boundary Perspective} 
\label{sec:limitation}
As discussed in Section \ref{sec:definition}, open-domain long-form generation relies on retrieved information to composite the article.
From a boundary perspective, redundancy can be analyzed in two aspects.
First, when the retrieved content contains \textit{limited factual knowledge}, the available information for generating the text is constrained, leading to redundancy in the generated article~\cite{lewis2021retrievalaugmentedgenerationknowledgeintensivenlp}.
Second, even when a large amount of non-redundant factual knowledge is retrieved, the model cannot organize and structure the knowledge as humans do to effectively utilize it, resulting in a limited amount of usable information and, consequently, redundancy~\cite{xia-etal-2024-rule}.
Similarly, the lack of novelty can be attributed to either the failure to collect novel knowledge or the inability to use the retrieved novel knowledge effectively.

In summary, the challenges in open-domain long-form generation can be abstracted into two knowledge boundary issues: the Knowledge Information Boundary and the Knowledge Cognition Boundary.

\begin{figure*}[htbp]
    \centering
    \includegraphics[width=0.9\textwidth]{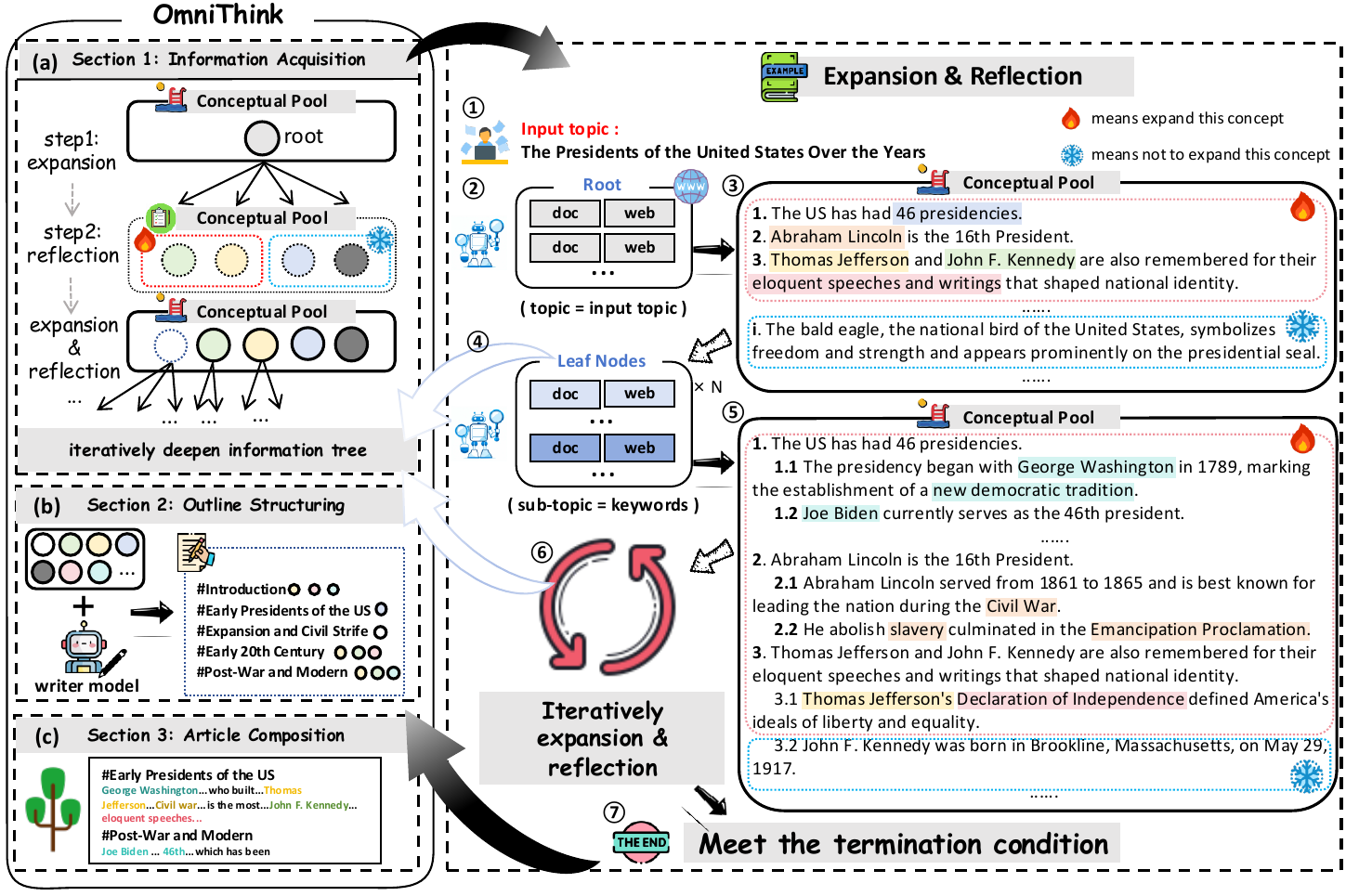}
    \caption{
        The overview of OmniThink. As shown in the left diagram, OmniThink is mainly divided into three steps:
        (a) Information Acquisition, (b) Outline Structuring, and (c) Article Composition. 
        The right diagram illustrates the specific operations during the Information Acquisition step. 
        (\textcircled{1} - \textcircled{2}) denotes the initialization of Information Acquisition, (\textcircled{2} - \textcircled{3}) corresponds to the reflection, and (\textcircled{3} - \textcircled{4} ) indicates the expansion.
    }
    \label{fig:main}
\end{figure*}

\begin{table}[t]
\centering
\resizebox{\columnwidth}{!}{
\begin{tabular}{lccc}
\hline
\textbf{Feature} & \textbf{STORM} & \textbf{Co-STORM} & \textbf{OmniThink} \\
\hline
Dynamic retrieval     & \ding{55} & \ding{55} & \checkmark \\
Structured memory          & \ding{55} & \checkmark & \checkmark \\
Reflective thinking   & \ding{55} & \ding{55} & \checkmark \\
\hline
\end{tabular}
}
\caption{Comparison of different methods. For more detailed explanations, please refer to the appendix \ref{appendix:compare}.}
\label{table:compare}
\end{table}

\section{OmniThink}
\label{sec:omnithink}

We introduce a machine writing framework OmniThink, which emulates the human slow-thinking process, as shown in Figure \ref{fig:main}.

% Iterative reflection on the retrieved content expands the Utilization Boundary, while iterative expansion based on the retrieved content extends the Knowledge Boundary.

%expansion和reflection是手段，而目的是为了得到information tree和conceptual pool。
%通过information tree和conceptual pool来expand the boundary
\subsection{Information Acquisition}
\label{sec:acquiringinformation}

% To acquire diverse and comprehensive information, OmniThink emulates the human slow-thinking process, progressively deepening its understanding of the topic through iterative \textbf{Expansion} and \textbf{Reflection}.
% As shown in Figure \ref{fig:main}, we illustrate the specific process of Expansion and Reflection. 
% This iterative process culminates in the construction of an information tree $\mathcal{T}$, which organizes the retrieved information in a structured and hierarchical manner, and a conceptual pool $\mathcal{P}$, which represents the LLMs' current understanding of the topic at time step $m$. 
% Together, these components form the foundation of article generation.

While LLMs have learned vast amounts of human knowledge through training, they may struggle to capture the spontaneous processes by which humans organize useful information and update cognitive frameworks when learning new knowledge~\cite{riva2024psychomaticsmultidisciplinaryframework,chemero2023llms}.
To address this, we propose two novel components: the \textbf{Information Tree $\mathcal{T}$} and the \textbf{Conceptual Pool $\mathcal{P}$} to simulate the human process of acquiring knowledge and updating cognitive frameworks~\cite{wu2025agenticreasoningreasoningllms}. 
Through interactive expansion and reflection, as shown in Figure \ref{fig:main}, these components are iteratively enriched, expanding the knowledge boundaries of open-domain long-form generation.

\paragraph{Initialization}
The interactive process begins with the initialization of a root node based on the input topic $\mathrm{T}$.
OmniThink first utilizes search engines, \textit{e.g.}, Google, or Bing, to retrieve information related to $\mathrm{T}$, using the retrieved information to construct the initial root node of the information tree $N_r$.
This initial information in $N_r$ is then analyzed and extracted to form a preliminary conceptual pool $\mathcal{P}_0$, which serves as OmniThink’s foundational cognition of the topic and guides subsequent expansion processes.

\subsubsection{Expansion of Information Tree}
\label{sec:expansion}
At time step $m$, OmniThink analyzes all leaf nodes $L_m = \{ N_0, N_1, \ldots, N_n \} $ of the information tree $ \mathcal{T}_m $. 
% These leaf nodes are first stored in the conceptual buffer $\mathcal{P}_b$, where each node is evaluated to determine if it requires further expansion. 
For nodes that need expansion, OmniThink uses the current conceptual pool $\mathcal{P}_m$ to identify areas for deeper expansion or suitable directions for expansion.
For each leaf node $N_i$, OmniThink generates $k_{N_i}$ sub-nodes, denoted as $\text{SUB}(N_i) = \{ S_0, S_1, \ldots, S_{k_{N_i}} \}$, for expansion. 
Each sub-node represents a specific aspect or subtopic identified from the current node $N_i$.
For each sub-node, OmniThink retrieves relevant information and stores it within the respective node, subsequently adding the sub-node to the appropriate position in the updated information tree $\mathcal{T}_{m+1}$ as follows:
\begin{equation} 
\mathcal{T}_{m+1} = \text{Combine}(\mathcal{T}_m, \text{SUB}(N_0), \ldots, \text{SUB}(N_n))
\end{equation} 

This targeted retrieval process ensures that OmniThink collects comprehensive and in-depth knowledge for each sub-node, thereby enriching the hierarchical structure of the information tree.

\subsubsection{Reflection of Conceptual Pool}
\label{sec:reflection}
In this phase, OmniThink reflects the newly retrieved information in all leaf nodes $L_{m+1} = \{ N_0,...N_n \}$ to update its cognitive framework, which is represented as conceptual pool.
The information from leaf nodes is analyzed, filtered, and synthesized to distill the core insights $I_{m+1} = \{ \text{INS}_0,..., \text{INS}_n \}$. 
These distilled insights are then incorporated into the conceptual pool $\mathcal{P}_m$, which is continuously updated and enriched throughout the process as follows:
\begin{equation}
    \mathcal{P}_{m+1} = \text{Merge}(I_{m+1}, \mathcal{P}_m)
\end{equation}
Using the updated conceptual pool $\mathcal{P}_{m+1}$, which represents the LLM's expanded cognition boundary on the topic, OmniThink further expands the leaf nodes of the information tree iteratively.

The iterative cycle of expansion and reflection continues until OmniThink determines that sufficient information has been acquired or the predefined maximum retrieval depth $K$ is reached. More details about the termination conditions can be found in Appendix \ref{appendix:decision}. 
During this process, as the Information Tree and Conceptual Pool are continuously expanded, the Information Boundary and Cognition Boundary are progressively expanded.
% The pseudocode for expansion and reflection can be found in Algorithm~\ref{alg:OmniThink}.

\subsection{Concept-guided  Outline Structuring}
\label{sec:creatingoutline}
%润一下
The outline determines the content direction, structural hierarchy, and logical progression of an article. 
To create an outline that is well-guided, clearly structured, and logically coherent, it is essential to have a comprehensive and in-depth cognition of the topic.
In the previous section, OmniThink maintains a conceptual pool that essentially represents the cognition boundary of the LLM.
When generating the content outline, we first create a draft outline $O_D$, and then ask the LLM to refine and link the content from the conceptual pool $\mathcal{P}$, ultimately forming the final outline $O = \text{Polish}(O_D, \mathcal{P})$.
Through this approach, the LLM is able to comprehensively cover the key points of the topic in the outline and ensure logical consistency and content coherence in the article.

\subsection{Article Composition}
\label{sec:writingarticle}

After completing the outline $O$, we begin writing for each section $S$. 
At this stage, the LLM would work in parallel for each section. 
When writing the content of the section, we use the titles of each section and their hierarchical subsections to retrieve the most relevant $K$ documents from the information tree by calculating the semantic similarity (Sentence-BERT \cite{reimers-gurevych-2019-sentence} embeddings).
After obtaining the relevant information, the LLM is prompted to generate the section content with citations based on the retrieved information.
Once all sections are generated, they will be concatenated into a complete draft article $\mathcal{A}_D = \{S_1,..S_n\}$. 
Since these sections are generated in parallel and the specific content of other sections is not yet clear, we prompt the LLM to process the concatenated article, remove redundant information, and form the final article $\mathcal{A} = \{S^{'}_1,..S^{'}_n\}$.

\section{Experiments}
\label{sec:setup}

\subsection{Dataset and Baseline}
We use WildSeek as evaluation dataset to verify the effectiveness of our method, following previous work~\cite{jiang2024unknownunknowns}.
WildSeek includes 100 data points across 24 different domains with each data consisting of a specific topic and a user's intend.
We select representative baselines for comparison, including RAG, oRAG, and STORM~\cite{shao2024assisting} and Co-STORM~\cite{jiang2024unknownunknowns}.
The baseline results are reproduced on the basis of STORM\footnote{\url{https://github.com/stanford-oval/storm}}.

\subsection{Knowledge Density Metric}
\label{sec:kd}
Previous works mostly focus on whether the article is relevant and correct, but do not consider whether the article is sufficiently concise and free of redundancy~\cite{li2024leveraginglargelanguagemodels,que2024hellobenchevaluatinglongtext,liu2024longgenbenchlongcontextgenerationbenchmark}. 
Many generated articles contain a lot of redundant information, which is very inconsistent with human writing.
To quantify this, we introduce the Knowledge Density (\textsc{KD}) for the generated article, which is defined as the ratio of meaningful content to the overall volume of text~\cite{said} as:
\begin{equation}
KD = \frac{\sum_{i=1}^{N}\mathcal{U}(k_i)}{L}
\end{equation}

where $N$ is the total number of atomic knowledge units identified within the document. 
The function $\mathcal{U}(k_i)$ indicates whether the $i$-th unit information $k_i$ is unique. 
$L$ represents the total length of the text. 
% In this formula, the numerator represents the sum of unique units of atomic knowledge extracted from a long article. 
% The denominator corresponds to the length of the article. 

In the appendix \ref{appendix:kd}, we empirically demonstrate the effectiveness of the KD metric.
% Note that the value of the knowledge density metric lies in its ability to measure the reading cost of generated text from the perspective of information acquisition~\cite{bovair1991toward,dos1993minimizing}.
Readers encountering low \textsc{KD} content often experience fatigue, frustration, or disengagement due to redundant or irrelevant details. 
In contrast, high-density content provides a streamlined experience, enabling efficient knowledge transfer.

% Previous methods exhibit limited performance on the proposed \textsc{KD} due to the fact that the generated content in open-domain long-form generation must be based on the retrieved information. 
% When the retrieved information is not sufficiently diverse, it often contains large amounts of redundant and repetitive content, leading to repetition and redundancy in the generated article. 
% This leaves room for optimizing the knowledge density in open-domain long-form generation. 
% We can address this issue by incorporating reasoning and planning during Step \textit{(i)}, where we process the gathered content to extract non-overlapping, high-density information \cite{qiao-etal-2023-reasoning,zelikman2024quietstarlanguagemodelsteach}.
\subsection{Evaluation Setup}
% We employ both automatic and human evaluations to assess the generated long-form articles:
We use Prometheus2~\cite{kim2024prometheus2opensource}\footnote{\url{https://github.com/prometheus-eval/prometheus-eval}} to automaticly score articles on a scale of 0 to 5, evaluating Relevance, Breadth, Depth, and Novelty. 
Furthermore, we measure information diversity~\cite{jiang2024unknownunknowns} (cosine similarity differences between web pages) and knowledge density (discussed in detail in \S\ref{sec:kd}) for information richness. 
Detailed procedures are provided in the Appendix~\ref{appendix:autoeval}. 
In addition, we also conduct a detailed human evaluation. 
The implementation details and evaluation results can be found in Appendix \ref{appendix:huamneval}.

\begin{table*}[t]
\centering
\small
\resizebox{0.95\textwidth}{!}{%
\begin{tabular}{clcccc!{\vrule width \lightrulewidth}c|c} 
\toprule
\multirow{2}{*}{\textbf{\makecell[c]{Backbones}}}& \multirow{2}{*}{\textbf{Methods}} & \multicolumn{4}{c!{\vrule width \lightrulewidth}}{\textbf{Rubric Grading}} & \multirow{2}{*}{\textbf{Information Diversity}} & \multirow{2}{*}{\textbf{Knowledge Density}} \\
            \cmidrule{3-6}
            &            & \multicolumn{1}{c}{Relevance}& \multicolumn{1}{c}{Breadth} & \multicolumn{1}{c}{Depth} & Novelty &  &       \\

\cmidrule{1-8}
\multicolumn{8}{c}{\cellcolor{uclablue} \textbf{\textit{Conversational Models}}} \\
\cmidrule{1-8}

\multirow{6}{*}{\textbf{GPT-4o}}
            & RAG                     & 4.65     & 4.55    & 4.59     & 4.22     & 0.1042   & 22.11                          \\
            & oRAG                    & 2.38     & 3.63    & 2.56     & 2.27     & 0.0963  & 19.70                           \\
            & STORM                   & 4.34     & 4.21    & 4.21     & 3.80     & 0.6342  & 19.33                          \\ 
             & Co-STORM*              & 4.37     & 4.66    & 4.65     & 3.89     & 0.6285  & 19.53                           \\
            & \cellcolor{mygray}{OmniThink}              & \cellcolor{mygray}{\textbf{4.77}}     & \cellcolor{mygray}{\textbf{4.71}}    & \cellcolor{mygray}{\textbf{4.66}}     & \cellcolor{mygray}{\textbf{4.31}}     & \cellcolor{mygray}{\textbf{0.6642}}  & \cellcolor{mygray}{\textbf{22.31}}  
\\
\cmidrule{1-8}
\multirow{6}{*}{\textbf{Qwen-Plus}}
            & RAG                     & 2.63     & 2.82    & 2.93     & 2.21     & 0.0927   & 10.32                          \\
            & oRAG                   & 2.42     & 2.52    & 2.66     & 2.22     & 0.1032  & 11.31                           \\
            & STORM                  & 2.72     & 2.81    & 3.00     & 2.72     & 0.6417  & 10.28                          \\ 
            & Co-STORM*              & 3.26     & 3.10    & 3.07     & 2.73     & 0.5332  & 11.52                          \\
            & \cellcolor{mygray}{OmniThink}              & \cellcolor{mygray}{\textbf{4.00}}     & \cellcolor{mygray}{\textbf{3.92}}    & \cellcolor{mygray}{\textbf{4.06}}     & \cellcolor{mygray}{\textbf{3.38}}     & \cellcolor{mygray}{\textbf{0.7230}}  & \cellcolor{mygray}{\textbf{11.66}}  
\\
\cmidrule{1-8}
\multicolumn{8}{c}{\cellcolor{uclagold} \textbf{\textit{Reasoning Models}}} \\
\cmidrule{1-8}
\multirow{6}{*}{\textbf{O1-preview}}
            & RAG                    & 3.99     & 4.13    & 4.02     & 3.44     & 0.1065   & 10.49                          \\
            & oRAG                   & 2.49     & 3.03    & 2.89     & 2.55     & 0.1222   & 10.51                     \\
            & STORM                  & 3.26     & 3.22    & 3.44     & 2.56     & 0.6121   & 10.82                             \\ 
            & Co-STORM*              & 3.41     & 3.29    & 3.23     & 2.97     & 0.6347   & 10.33                               \\
            & \cellcolor{mygray}{OmniThink}              & \cellcolor{mygray}{\textbf{4.20}}     & \cellcolor{mygray}{\textbf{4.20}}    & \cellcolor{mygray}{\textbf{4.32}}     & \cellcolor{mygray}{\textbf{3.60}}     & \cellcolor{mygray}{\textbf{0.6752}}  & \cellcolor{mygray}{\textbf{10.87}}  
\\
\cmidrule{1-8}

\multirow{6}{*}{\textbf{DeepSeek-R1}}
            & RAG                    & 4.12     & 4.33    & 4.55     & 4.44     & 0.1044   & 11.32                          \\
            & oRAG                   & 4.56     & 4.49    & 4.39     & 4.37     & 0.1123   & 10.44                     \\
            & STORM                  & 2.42     & 2.93    & 3.14     & 2.86     & 0.6640   & 11.57                             \\ 
            & Co-STORM*              & 4.62     & 4.54    & 4.78     & 4.47     & 0.5332   & 11.66                               \\
            & \cellcolor{mygray}{OmniThink}              & \cellcolor{mygray}{\textbf{4.70}}     & \cellcolor{mygray}{\textbf{4.78}}    & \cellcolor{mygray}{\textbf{4.78}}     & \cellcolor{mygray}{\textbf{4.59}}     & \cellcolor{mygray}{\textbf{0.6653}}  & \cellcolor{mygray}{\textbf{11.72}}  
\\

\bottomrule
\end{tabular}
}

\caption{Results of article quality evaluation.
        $^*$ means that this method is different from the original experimental setting, primarily in the human-machine collaboration component. 
        Instead of simulating human involvement through an agent, as done in the original paper~\cite{jiang2024unknownunknowns}, we remove the human participation step. The variance of evaluation can be found in Appendix \ref{appendix:variance}.}
\label{tab:main}
\end{table*}

\subsection{Implementation Details}
We build OmniThink based on the DSpy framework~\cite{khattab2023dspycompilingdeclarativelanguage}, and Appendix~\ref{appendix:fullprompt} contains the corresponding prompts we used. 
During generation, we set the \textit{temperature} at 1.0 and \textit{top\_p} at 0.9. 
We use Bing's API with the parameter for the number of web pages returned per query set to 5.
%To retrieve information based on the outline, we use SentenceBERT~\cite{reimers-gurevych-2019-sentence} embeddings to calculate cosine similarity thereby retrieving the three most similar web pages each time.
For the computation of knowledge density, we utilize Factscore\footnote{\url{https://github.com/shmsw25/FActScore}} with \texttt{GPT-4o-08-06} as the backbone to decompose atomic knowledge ~\cite{min2023factscorefinegrainedatomicevaluation}. 
After decomposition, we proceed to use GPT-4o-08-06 for the deduplication of the split atomic knowledge.
To avoid the impact of search engine changes over time.
More implementation details are presented in Appendix~\ref{app:implementation}.

\subsection{Main Results}
%先讲清楚boundary，再讲清楚我们扩大了boundary是不是真的高效利用了

\paragraph{Article Generation.}
Table~\ref{tab:main} presents the evaluation results on WildSeek dataset.
Within the framework of four grading criteria (Relevance, Breadth, Depth, and Novelty) OmniThink excels across all metrics, particularly standing out in Novelty.
This achievement can be attributed to OmniThink's Information Tree and Conceptual Pool, which are continuously enriched, enabling OmniThink to expand the boundaries of existing knowledge.
% Section~\ref{sec:er} presents a detailed discussion of these results.

OmniThink utilizes the Conceptual Pool for multidimensional deep thinking on the retrieved information during the retrieval process, enabling subsequent searches to access deeper levels of external knowledge, thereby enhancing the diversity of information.

In terms of knowledge density, OmniThink employs a continuous and dynamic retrieval strategy, storing a wealth of information in the Information Tree. 
This allows OmniThink to draw upon a broader range of resources during the content generation phase, positioning OmniThink at a distinct advantage in the knowledge density metric compared to existing benchmark methods.

\paragraph{Outline Generation.}
%原因 结果 分析
\begin{table}[h]
\centering
\resizebox{\columnwidth}{!}{%
\begin{tabular}{lccc} 
\toprule
\textbf{Method}                          & \makecell{\textbf{Content}\\\textbf{Guidance}}   &  \makecell{\textbf{Hierarchical}\\\textbf{Clarity}}           & \makecell{\textbf{Logical}\\\textbf{Coherence}}                       \\ 
\midrule
oRAG                            & 3.93       & 3.95              & 3.97                            \\
STORM                           & 3.92       & 3.99              & 3.99                            \\
Co-STORM*                       & 3.45       & 3.27              & 3.41                            \\
\cellcolor{mygray}{OmniThink}   & \cellcolor{mygray}{\textbf{4.00}}     & \cellcolor{mygray}{\textbf{4.02}}              & \cellcolor{mygray}{\textbf{3.99}}                            \\

\bottomrule
\end{tabular}
}
\caption{Results of outline quality evaluation.}
\label{table:outline_quality}
\vspace{-0.7em}
\end{table}

% The outline serves as a critical intermediary in the process of article generation, with its quality exerting a direct impact on the coherence, logical consistency, and expressive clarity of the final article. 
%For articles without a specific stylistic constraint (\textit{e.g.}, Wikipedia-like), it may be unnecessary to overemphasize the entity recall rate of the outline. 
%Emphasis should be placed on its hierarchical clarity, logical coherence, and its ability to effectively guide the content generation process.
We evaluate outline quality from the perspectives of structural soundness, logical consistency, and generative guidance. 
More evaluation details can be found in the Appendix \ref{appendix:autoeval_outline}.
From Table \ref{table:outline_quality}, we notice that OmniThink achieves superior performance. 
This improvement can be attributed to the unique design of OmniThink's Conceptual Pool, which enables the LLMs to develop a more comprehensive and diverse understanding of the target topic during outline generation. 
% Consequently, this facilitates better guidance for content production and enhances the overall structural coherence of the generated content. 

\section{Analysis}
\subsection{Ablation Study}
\paragraph{Information tree and Conceptual pool Ablation.}
For the Information Tree, we remove the hierarchical structure and instead have the OmniThink reflect over all retrieved content directly, followed by another retrieval. 
In contrast, to evaluate the Conceptual Pool, we disable reflection and allow the Information Tree to grow continuously until the maximum depth of Information tree is reached.
As shown in Figure \ref{fig:er}(a) and Figure \ref{fig:er}(b), the performance of OmniThink degrades when either the Information Tree or the Conceptual Pool is removed.

\paragraph{Expansion and Reflection Ablation.}
We compare OmniThink with a version that does not implement expansion and reflection. 
As shown in Figure \ref{fig:er}(c), w/o E\&R performs worse in all metrics than the complete system, particularly in terms of Information Diversity and Novelty. 

\subsection{Boundary Analysis}
\label{sec:boundaryanalysis}
\begin{figure}[h]
    \centering
    \includegraphics[width=0.85\linewidth]{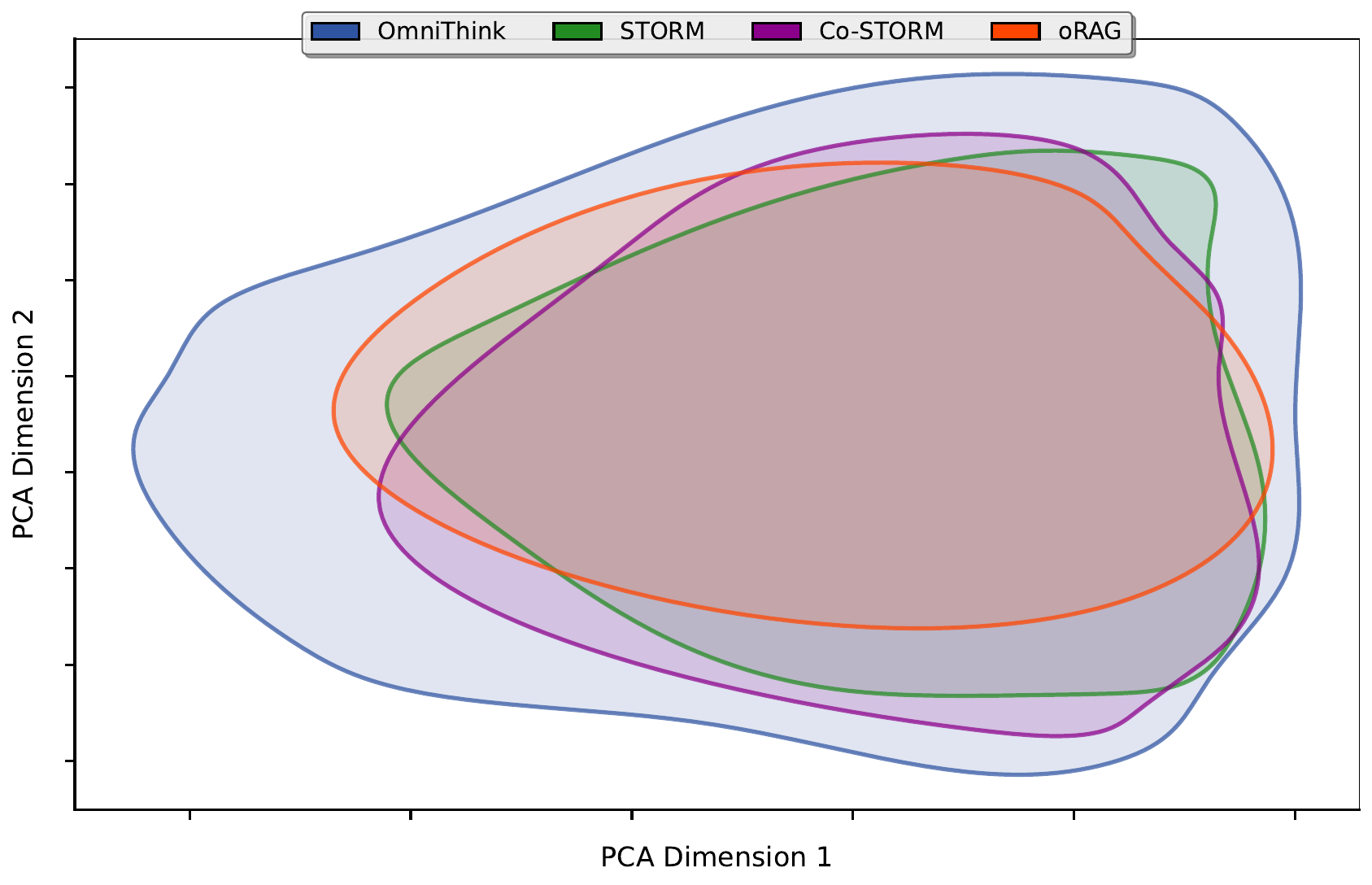}
    \caption{
    The information scope of OmniThink, Co-STORM, STORM and oRAG.
}
    \label{fig:knowledgeboundary}
\end{figure}

As discussed in Section \ref{sec:limitation}, we divide the boundary into Information Boundary and Cognition Boundary. 
In this section, we explore in detail whether OmniThink has truly expanded these boundaries.
\paragraph{Information Boundary.}
To investigate whether OmniThink has truly expanded the Information Boundary, we map the retrieval information of OmniThink, STORM, and Co-STORM to a two-dimensional plane as their Information Boundary to visualize the scope.
As shown in Figure \ref{fig:knowledgeboundary}, OmniThink has the largest retrieval scope, indicating that it has indeed expanded the Information Boundary through the information tree and conceptual pool.
More implementation details can be found in Appendix \ref{appendix:informationboundary}.

%脱离了结构化知识，导致效果降低
\paragraph{Cognition Boundary.}
\begin{figure}[t]
    \centering
    \includegraphics[width=0.85\linewidth]{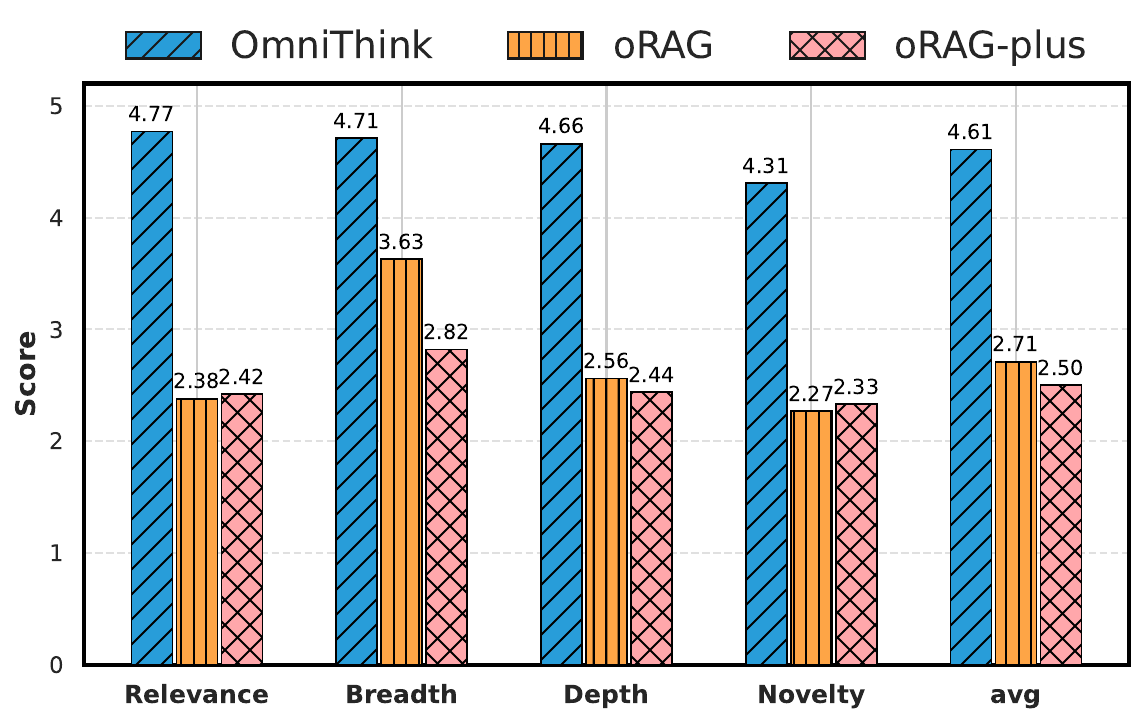}
    \caption{
    The Comparison of results between OmniThink, oRAG, and oRAG-plus.
}
    \label{fig:knowledgecognition}
\end{figure}
For the Cognition Boundary, since Expansion and Reflection cannot be separated, we set a new baseline, oRAG-Plus, where we increase the number of web pages retrieved by oRAG-Plus to match that of OmniThink. 
% The results are shown in the Table \ref{fig:cognitionboundary}. 
From Figure~\ref{fig:knowledgecognition}, it can be observed that without the guidance of the Conceptual Pool, even with a large amount of information, the LLM still fails to utilize it effectively. 
In fact, some of the results of oRAG-Plus are even lower than those of oRAG, which may be due to the lack of sufficient cognition to utilize the retrieved information, with excessive web content acting as noise to the model.

\subsection{Expansion \& Reflection Analysis}
\label{sec:er}

\begin{figure*}[t]
    \centering
    \includegraphics[width=0.85\linewidth]{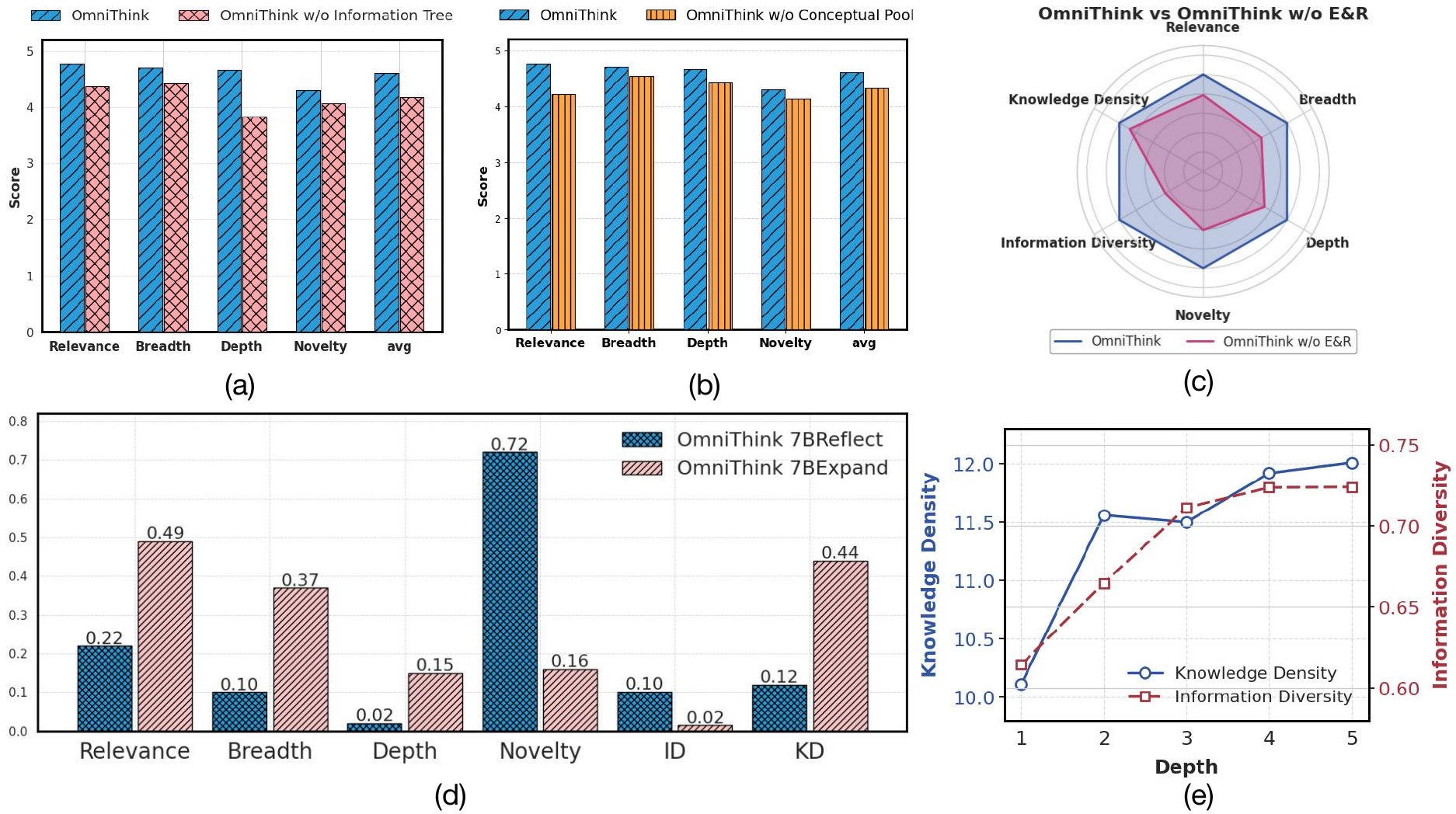}
    \caption{
    (a) The Ablation of Conceptual Pool; (b) The Ablation of Information Tree; 
    (c) The Ablation of OmniThink, OmniThink w/o E\&R represents a version of OmniThink without expansion and reflection
    ; (d) The comparison of the impact of expansion and reflection on various metrics, OmniThink 7BReflect indicates the use of Qwen2.5-7b-instruct for Reflection. More details can be found in Appendix \ref{appendix:er}
    ; (e) The result of depth analysis.
}
    \label{fig:er}
\end{figure*}
% In this section, we provide a detailed analysis of the expansion and reflection.
% This experiment demonstrates the powerful role of the dynamic expansion and reflection mechanism in enhancing information diversity and article novelty.
%In the previous chapters, we discussed the effectiveness of Expansion and Reflection within OmniThink.
%Building upon this foundation, 

\paragraph{Cognitive boundary mainly constrain the potential for innovation.}
To further analyze how the expansion and reflection processes shape various aspects of the final article through the conceptual pool and information tree, we design an indirect yet ingenious experiment. 
As shown in Figure \ref{fig:er}(b), we use lower-performing models to complete the expansion and reflection processes, with the decline in various metrics serving as an indicator of their impact on the article. 
The details of the experimental design can be found in Appendix \ref{appendix:er}.
% We report the results in Figure \ref{fig:er}(b).
We observe that reflection is much more important for novelty.
As discussed in Section \ref{sec:boundaryanalysis}, OmniThink indeed expands the knowledge boundary.
Reflection endows the model with the ability not only to re-evaluate and introspectively consider existing knowledge but also to integrate this information in a way that promotes the emergence of more diverse and expansive ideas, which is similar to our definition of the cognition boundary.
Expanding the cognition boundary through Reflection significantly enhances the model's innovation in generating articles.
Therefore, we believe that it is the cognition boundary that limits the model's writing innovation.

\paragraph{Information boundary limits the effective organization of information on the topic.}
We notice that expansion is more important than reflection in Knowledge Density, Breadth, and Depth.
The rationale behind this is that expansion inherently sets the trajectory for the model's subsequent information retrieval.
By establishing more precise and effective directions for the model's retrieval process, it becomes more adept at harnessing the retrieved information to expand the information boundary. 
This integration not only enhances the relevance of the content but also increases the knowledge density, as the text becomes more comprehensive and nuanced. 
Consequently, a better expansion strategy leads to a more sophisticated planner, capable of navigating the complexities of information retrieval and utilization with greater finesse.

\paragraph{More knowledge boundaries need to be identified and defined.}

Previous experiments have shown that expansion and reflection extend the information boundary and cognition boundary, which improves the quality of the articles. 
We increase the depth of expansion and reflection to explore how far they can extend the knowledge boundary.
From Figure \ref{fig:er}(c), we observe that as the depth increases, the growth rate of knowledge density and information diversity significantly slows down. 
This indicates that the information boundary and cognition boundary are no longer the primary limitations on article quality, and other boundaries need to be identified and defined.

\section{Related Work}
\subsection{Information Seeking in NLP}
Previous studies on information-seeking focused on designing question-answering (QA) systems~\cite{wu2025webwalker}. 
Early open-domain QA methods generally assumed that users could fulfill their information needs through a single query ~\cite{chen2017readingwikipediaansweropendomain, levy2021opendomainquestionansweringcovid19emergent}.
Subsequent studies have recognized that, in real-world scenarios, users often struggle to satisfy their information needs with a single query \cite{chen2017readingwikipediaansweropendomain, levy2021opendomainquestionansweringcovid19emergent}. 
To address this limitation, researchers have explored \textit{multi sub-query} retrieval methods, where a single query is decomposed into multiple sub-queries to retrieve distinct pieces of information~\cite{mao-etal-2024-rafe,chen2011optimization,peng2019optimizing}. 
The information collected is then aggregated to provide a comprehensive answer.
Building on these developments, recent advances in open-domain long-form generation require reasoning across multiple information sources~\cite{fan2019eli5,ujwal2024reasoning,wei2024long,tan2024proxyqaalternativeframeworkevaluating}. 
This line of open-domain long-form generatio underscores the importance of integrating information from multiple perspectives. 
% For example, STORM introduces a retrieval paradigm that simulates multi-turn interactions from diverse perspectives, aiming to aggregate richer and more diverse information\cite{shao2024collaborativegymframeworkenabling}. 
% Similarly, Co-STORM employs a ``roundtable discussion'' paradigm to further expand the diversity of information sources considered during retrieval~\cite{jiang2024unknownunknowns}. 
% Although these approaches have made significant advancements from \textit{multi-perspective} and \textit{multi sub-query} perspectives, they often fail to leverage the reasoning and introspective abilities of LLMs fully. 
% Specifically, existing approaches do not fully exploit the potential of LLMs to dynamically adjust retrieval strategies and flexibly update information sources as the model's understanding of the topic deepens\cite{qin2024o1replicationjourneystrategic}.
% Unlike previous methods, we propose an approach that simulates the human slow-thinking process, where the model synthesizes and updates its cognitive framework based on existing retrieval information to further retrieve additional information.

\subsection{Machine Writing}
Due to the high costs associated with manual writing, machine writing has garnered significant research interest in recent years~\cite{zhou2023recurrentgptinteractivegenerationarbitrarily,pham2024suri,wang2024autopatentmultiagentframeworkautomatic,wang2024weaver, 2024autosurvey}.
The emergence of LLMs and Retrieval-Augmented Generation (RAG) has opened new possibilities for automated writing~\cite{liang2024integratingplanningsingleturnlongform,balepur2023expositorytextgenerationimitate,de_la_Torre_L_pez_2023}. 
To ensure authenticity and real-time relevance, current RAG-based automated writing systems primarily rely on retrieved content to generate articles. 
For example, STORM~\cite{shao2024assisting} introduces a role-playing question-and-answer approach to author Wikipedia-like articles, while Co-STORM~\cite{jiang2024unknownunknowns} proposes a user-participated information retrieval paradigm. 
% Besides, AutoSurvey~\cite{2024autosurvey} extends this framework into the domain of academic paper writing. 
% However, these methods tend to overlook the issue of information diversity, which can result in outputs with limited practical value.
% Although these methods demonstrate notable advancements in specific domains, they often neglect the perspective of content utility, resulting in outputs with limited practical value.
% We propose a new strategy, which starts from the perspective of information sources, providing LLMs with higher quality knowledge to subsequently improve the quality of the generated text.

\section{Conclusion and Future Work}
We propose OmniThink, a machine writing framework that emulates the human-like process of iterative expansion and reflection. 
Automatic and human evaluations demonstrate that OmniThink can generate well-founded, high-quality long articles. 
OmniThink is model-agnostic and can be integrated with existing frameworks. 
In the future, we will explore more advanced machine writing methods that combine deeper reasoning with human-computer interaction.

\newpage
\section*{Limitations}
%We design OmniThink, a writing  with continuous expansion and reflection.
Although the proposed OmniThink has demonstrated its advantages in both automatic and human evaluations, several limitations remain.
Firstly, the current work is limited to search and text generation, while a vast amount of multimodal information in the open domain remains unused. 
Secondly, we have not considered personalized language styles in text production. 
As a result, the generated texts tend to be academic in nature, which may not be as suitable for general users' reading preferences.
We plan to address these limitations in future work.

\section*{Acknowledgements}
This work was supported by the National Natural Science Foundation of China (No. 62576307, No. NSFCU23B2055, No. NSFCU19B2027), the Fundamental Research Funds for the Central Universities (226-2023-00138), Yongjiang Talent Introduction Programme (2021A-156-G), Ningbo Natural Science Foundation (2024J020), Information Technology Center and State Key Lab of CAD\&CG, Zhejiang University.

\bibliography{custom}

\begin{thebibliography}{54}
\providecommand{\natexlab}[1]{#1}

\bibitem[{Balepur et~al.(2023)Balepur, Huang, and Chang}]{balepur2023expositorytextgenerationimitate}
Nishant Balepur, Jie Huang, and Kevin Chen-Chuan Chang. 2023.
\newblock \href {https://arxiv.org/abs/2305.03276} {Expository text generation: Imitate, retrieve, paraphrase}.
\newblock \emph{Preprint}, arXiv:2305.03276.

\bibitem[{Bean and Melzer(2021)}]{bean2021engaging}
John~C Bean and Dan Melzer. 2021.
\newblock \emph{Engaging ideas: The professor's guide to integrating writing, critical thinking, and active learning in the classroom}.
\newblock John Wiley \& Sons.

\bibitem[{Bruce(1978)}]{bruce1978cognitive}
Bertram~C Bruce. 1978.
\newblock A cognitive science approach to writing.
\newblock \emph{Center for the Study of Reading Technical Report; no. 089}.

\bibitem[{Chemero(2023)}]{chemero2023llms}
Anthony Chemero. 2023.
\newblock Llms differ from human cognition because they are not embodied.
\newblock \emph{Nature Human Behaviour}, 7(11):1828--1829.

\bibitem[{Chen et~al.(2017)Chen, Fisch, Weston, and Bordes}]{chen2017readingwikipediaansweropendomain}
Danqi Chen, Adam Fisch, Jason Weston, and Antoine Bordes. 2017.
\newblock \href {https://arxiv.org/abs/1704.00051} {Reading wikipedia to answer open-domain questions}.
\newblock \emph{Preprint}, arXiv:1704.00051.

\bibitem[{Chen et~al.(2011)Chen, Wu, Liu, Yang, and Zheng}]{chen2011optimization}
Gang Chen, Yongwei Wu, Jia Liu, Guangwen Yang, and Weimin Zheng. 2011.
\newblock Optimization of sub-query processing in distributed data integration systems.
\newblock \emph{Journal of Network and Computer Applications}, 34(4):1035--1042.

\bibitem[{de~la Torre-López et~al.(2023)de~la Torre-López, Ramírez, and Romero}]{de_la_Torre_L_pez_2023}
José de~la Torre-López, Aurora Ramírez, and José~Raúl Romero. 2023.
\newblock \href {https://doi.org/10.1007/s00607-023-01181-x} {Artificial intelligence to automate the systematic review of scientific literature}.
\newblock \emph{Computing}, 105(10):2171–2194.

\bibitem[{Edge et~al.(2024)Edge, Trinh, Cheng, Bradley, Chao, Mody, Truitt, and Larson}]{edge2024local}
Darren Edge, Ha~Trinh, Newman Cheng, Joshua Bradley, Alex Chao, Apurva Mody, Steven Truitt, and Jonathan Larson. 2024.
\newblock From local to global: A graph rag approach to query-focused summarization.
\newblock \emph{arXiv preprint arXiv:2404.16130}.

\bibitem[{Fan et~al.(2019)Fan, Jernite, Perez, Grangier, Weston, and Auli}]{fan2019eli5}
Angela Fan, Yacine Jernite, Ethan Perez, David Grangier, Jason Weston, and Michael Auli. 2019.
\newblock Eli5: Long form question answering.
\newblock \emph{arXiv preprint arXiv:1907.09190}.

\bibitem[{Gao et~al.(2024)Gao, Xiong, Gao, Jia, Pan, Bi, Dai, Sun, Wang, and Wang}]{gao2024retrievalaugmentedgenerationlargelanguage}
Yunfan Gao, Yun Xiong, Xinyu Gao, Kangxiang Jia, Jinliu Pan, Yuxi Bi, Yi~Dai, Jiawei Sun, Meng Wang, and Haofen Wang. 2024.
\newblock \href {https://arxiv.org/abs/2312.10997} {Retrieval-augmented generation for large language models: A survey}.
\newblock \emph{Preprint}, arXiv:2312.10997.

\bibitem[{Ji et~al.(2025)Ji, Li, Ye, Wu, Xu, Mo, and Zhang}]{ji2025testtimecomputingsystem1thinking}
Yixin Ji, Juntao Li, Hai Ye, Kaixin Wu, Jia Xu, Linjian Mo, and Min Zhang. 2025.
\newblock \href {https://arxiv.org/abs/2501.02497} {Test-time computing: from system-1 thinking to system-2 thinking}.
\newblock \emph{Preprint}, arXiv:2501.02497.

\bibitem[{Jiang et~al.(2024)Jiang, Shao, Ma, Semnani, and Lam}]{jiang2024unknownunknowns}
Yucheng Jiang, Yijia Shao, Dekun Ma, Sina~J. Semnani, and Monica~S. Lam. 2024.
\newblock \href {https://arxiv.org/abs/2408.15232} {Into the unknown unknowns: Engaged human learning through participation in language model agent conversations}.
\newblock \emph{Preprint}, arXiv:2408.15232.

\bibitem[{Khattab et~al.(2023)Khattab, Singhvi, Maheshwari, Zhang, Santhanam, Vardhamanan, Haq, Sharma, Joshi, Moazam, Miller, Zaharia, and Potts}]{khattab2023dspycompilingdeclarativelanguage}
Omar Khattab, Arnav Singhvi, Paridhi Maheshwari, Zhiyuan Zhang, Keshav Santhanam, Sri Vardhamanan, Saiful Haq, Ashutosh Sharma, Thomas~T. Joshi, Hanna Moazam, Heather Miller, Matei Zaharia, and Christopher Potts. 2023.
\newblock \href {https://arxiv.org/abs/2310.03714} {Dspy: Compiling declarative language model calls into self-improving pipelines}.
\newblock \emph{Preprint}, arXiv:2310.03714.

\bibitem[{Kim et~al.(2024)Kim, Suk, Longpre, Lin, Shin, Welleck, Neubig, Lee, Lee, and Seo}]{kim2024prometheus2opensource}
Seungone Kim, Juyoung Suk, Shayne Longpre, Bill~Yuchen Lin, Jamin Shin, Sean Welleck, Graham Neubig, Moontae Lee, Kyungjae Lee, and Minjoon Seo. 2024.
\newblock \href {https://arxiv.org/abs/2405.01535} {Prometheus 2: An open source language model specialized in evaluating other language models}.
\newblock \emph{Preprint}, arXiv:2405.01535.

\bibitem[{Levy et~al.(2021)Levy, Mo, Xiong, and Wang}]{levy2021opendomainquestionansweringcovid19emergent}
Sharon Levy, Kevin Mo, Wenhan Xiong, and William~Yang Wang. 2021.
\newblock \href {https://arxiv.org/abs/2110.06962} {Open-domain question-answering for covid-19 and other emergent domains}.
\newblock \emph{Preprint}, arXiv:2110.06962.

\bibitem[{Lewis et~al.(2021)Lewis, Perez, Piktus, Petroni, Karpukhin, Goyal, Küttler, Lewis, tau Yih, Rocktäschel, Riedel, and Kiela}]{lewis2021retrievalaugmentedgenerationknowledgeintensivenlp}
Patrick Lewis, Ethan Perez, Aleksandra Piktus, Fabio Petroni, Vladimir Karpukhin, Naman Goyal, Heinrich Küttler, Mike Lewis, Wen tau Yih, Tim Rocktäschel, Sebastian Riedel, and Douwe Kiela. 2021.
\newblock \href {https://arxiv.org/abs/2005.11401} {Retrieval-augmented generation for knowledge-intensive nlp tasks}.
\newblock \emph{Preprint}, arXiv:2005.11401.

\bibitem[{Li et~al.(2024)Li, Xu, Shen, Xu, Gu, Lai, Tao, and Ma}]{li2024leveraginglargelanguagemodels}
Zhen Li, Xiaohan Xu, Tao Shen, Can Xu, Jia-Chen Gu, Yuxuan Lai, Chongyang Tao, and Shuai Ma. 2024.
\newblock \href {https://arxiv.org/abs/2401.07103} {Leveraging large language models for nlg evaluation: Advances and challenges}.
\newblock \emph{Preprint}, arXiv:2401.07103.

\bibitem[{Liang et~al.(2023)Liang, Tang, Li, and Zhang}]{liang-etal-2023-open}
Xiaobo Liang, Zecheng Tang, Juntao Li, and Min Zhang. 2023.
\newblock \href {https://doi.org/10.18653/v1/2023.acl-long.13} {Open-ended long text generation via masked language modeling}.
\newblock In \emph{Proceedings of the 61st Annual Meeting of the Association for Computational Linguistics (Volume 1: Long Papers)}, pages 223--241, Toronto, Canada. Association for Computational Linguistics.

\bibitem[{Liang et~al.(2024)Liang, Wu, Zhuang, Chen, Shen, Jia, Qin, Sanghai, Wang, Yang, and Bendersky}]{liang2024integratingplanningsingleturnlongform}
Yi~Liang, You Wu, Honglei Zhuang, Li~Chen, Jiaming Shen, Yiling Jia, Zhen Qin, Sumit Sanghai, Xuanhui Wang, Carl Yang, and Michael Bendersky. 2024.
\newblock \href {https://arxiv.org/abs/2410.06203} {Integrating planning into single-turn long-form text generation}.
\newblock \emph{Preprint}, arXiv:2410.06203.

\bibitem[{Liu et~al.(2018)Liu, Saleh, Pot, Goodrich, Sepassi, Kaiser, and Shazeer}]{liu2018generatingwikipediasummarizinglong}
Peter~J. Liu, Mohammad Saleh, Etienne Pot, Ben Goodrich, Ryan Sepassi, Lukasz Kaiser, and Noam Shazeer. 2018.
\newblock \href {https://arxiv.org/abs/1801.10198} {Generating wikipedia by summarizing long sequences}.
\newblock \emph{Preprint}, arXiv:1801.10198.

\bibitem[{Liu et~al.(2024)Liu, Dong, Hu, and Chu}]{liu2024longgenbenchlongcontextgenerationbenchmark}
Xiang Liu, Peijie Dong, Xuming Hu, and Xiaowen Chu. 2024.
\newblock \href {https://arxiv.org/abs/2410.04199} {Longgenbench: Long-context generation benchmark}.
\newblock \emph{Preprint}, arXiv:2410.04199.

\bibitem[{Mao et~al.(2024)Mao, Jiang, Chen, Li, Wang, Wang, Xie, Huang, Chen, and Zhang}]{mao-etal-2024-rafe}
Shengyu Mao, Yong Jiang, Boli Chen, Xiao Li, Peng Wang, Xinyu Wang, Pengjun Xie, Fei Huang, Huajun Chen, and Ningyu Zhang. 2024.
\newblock \href {https://doi.org/10.18653/v1/2024.findings-emnlp.49} {{R}a{F}e: Ranking feedback improves query rewriting for {RAG}}.
\newblock In \emph{Findings of the Association for Computational Linguistics: EMNLP 2024}, pages 884--901, Miami, Florida, USA. Association for Computational Linguistics.

\bibitem[{Min et~al.(2023)Min, Krishna, Lyu, Lewis, tau Yih, Koh, Iyyer, Zettlemoyer, and Hajishirzi}]{min2023factscorefinegrainedatomicevaluation}
Sewon Min, Kalpesh Krishna, Xinxi Lyu, Mike Lewis, Wen tau Yih, Pang~Wei Koh, Mohit Iyyer, Luke Zettlemoyer, and Hannaneh Hajishirzi. 2023.
\newblock \href {https://arxiv.org/abs/2305.14251} {Factscore: Fine-grained atomic evaluation of factual precision in long form text generation}.
\newblock \emph{Preprint}, arXiv:2305.14251.

\bibitem[{Osterman(1990)}]{osterman1990reflective}
Karen~F Osterman. 1990.
\newblock Reflective practice: A new agenda for education.
\newblock \emph{Education and urban society}, 22(2):133--152.

\bibitem[{Parmar et~al.(2010)Parmar, Freeman, Harrison, Wicks, Purnell, and De~Colle}]{parmar2010stakeholder}
Bidhan~L Parmar, R~Edward Freeman, Jeffrey~S Harrison, Andrew~C Wicks, Lauren Purnell, and Simone De~Colle. 2010.
\newblock Stakeholder theory: The state of the art.
\newblock \emph{Academy of Management Annals}, 4(1):403--445.

\bibitem[{Peng et~al.(2019)Peng, Ge, Zou, {\"O}zsu, Xu, and Zhao}]{peng2019optimizing}
Peng Peng, Qi~Ge, Lei Zou, M~Tamer {\"O}zsu, Zhiwei Xu, and Dongyan Zhao. 2019.
\newblock Optimizing multi-query evaluation in federated rdf systems.
\newblock \emph{IEEE Transactions on Knowledge and Data Engineering}, 33(4):1692--1707.

\bibitem[{Pham et~al.(2024)Pham, Sun, and Iyyer}]{pham2024suri}
Chau~Minh Pham, Simeng Sun, and Mohit Iyyer. 2024.
\newblock Suri: Multi-constraint instruction following for long-form text generation.
\newblock \emph{arXiv preprint arXiv:2406.19371}.

\bibitem[{Quan et~al.(2024)Quan, Tang, Yu, Yang, Liu, Gao, Tu, Zhang, Zhou, and Lin}]{quan2024languagemodelsselflengthengenerate}
Shanghaoran Quan, Tianyi Tang, Bowen Yu, An~Yang, Dayiheng Liu, Bofei Gao, Jianhong Tu, Yichang Zhang, Jingren Zhou, and Junyang Lin. 2024.
\newblock \href {https://arxiv.org/abs/2410.23933} {Language models can self-lengthen to generate long texts}.
\newblock \emph{Preprint}, arXiv:2410.23933.

\bibitem[{Que et~al.(2024)Que, Duan, He, Mou, Zhou, Liu, Rong, Wang, Yang, Zhang, Peng, Zhang, Zhang, and Chen}]{que2024hellobenchevaluatinglongtext}
Haoran Que, Feiyu Duan, Liqun He, Yutao Mou, Wangchunshu Zhou, Jiaheng Liu, Wenge Rong, Zekun~Moore Wang, Jian Yang, Ge~Zhang, Junran Peng, Zhaoxiang Zhang, Songyang Zhang, and Kai Chen. 2024.
\newblock \href {https://arxiv.org/abs/2409.16191} {Hellobench: Evaluating long text generation capabilities of large language models}.
\newblock \emph{Preprint}, arXiv:2409.16191.

\bibitem[{Ram et~al.(2023)Ram, Levine, Dalmedigos, Muhlgay, Shashua, Leyton-Brown, and Shoham}]{ram2023incontextretrievalaugmentedlanguagemodels}
Ori Ram, Yoav Levine, Itay Dalmedigos, Dor Muhlgay, Amnon Shashua, Kevin Leyton-Brown, and Yoav Shoham. 2023.
\newblock \href {https://arxiv.org/abs/2302.00083} {In-context retrieval-augmented language models}.
\newblock \emph{Preprint}, arXiv:2302.00083.

\bibitem[{Reimers and Gurevych(2019)}]{reimers-gurevych-2019-sentence}
Nils Reimers and Iryna Gurevych. 2019.
\newblock \href {https://doi.org/10.18653/v1/D19-1410} {Sentence-{BERT}: Sentence embeddings using {S}iamese {BERT}-networks}.
\newblock In \emph{Proceedings of the 2019 Conference on Empirical Methods in Natural Language Processing and the 9th International Joint Conference on Natural Language Processing (EMNLP-IJCNLP)}, pages 3982--3992, Hong Kong, China. Association for Computational Linguistics.

\bibitem[{Riva et~al.(2024)Riva, Mantovani, Wiederhold, Marchetti, and Gaggioli}]{riva2024psychomaticsmultidisciplinaryframework}
Giuseppe Riva, Fabrizia Mantovani, Brenda~K. Wiederhold, Antonella Marchetti, and Andrea Gaggioli. 2024.
\newblock \href {https://arxiv.org/abs/2407.16444} {Psychomatics -- a multidisciplinary framework for understanding artificial minds}.
\newblock \emph{Preprint}, arXiv:2407.16444.

\bibitem[{Shanahan et~al.(2023)Shanahan, McDonell, and Reynolds}]{shanahan2023role}
Murray Shanahan, Kyle McDonell, and Laria Reynolds. 2023.
\newblock Role play with large language models.
\newblock \emph{Nature}, 623(7987):493--498.

\bibitem[{Shao et~al.(2024)Shao, Jiang, Kanell, Xu, Khattab, and Lam}]{shao2024assisting}
Yijia Shao, Yucheng Jiang, Theodore~A. Kanell, Peter Xu, Omar Khattab, and Monica~S. Lam. 2024.
\newblock {Assisting in Writing Wikipedia-like Articles From Scratch with Large Language Models}.
\newblock In \emph{Proceedings of the 2024 Conference of the North American Chapter of the Association for Computational Linguistics: Human Language Technologies, Volume 1 (Long and Short Papers)}.

\bibitem[{Shen et~al.(2023)Shen, August, Siangliulue, Lo, Bragg, Hammerbacher, Downey, Chang, and Sontag}]{shen2023summarizationdesigningaisupport}
Zejiang Shen, Tal August, Pao Siangliulue, Kyle Lo, Jonathan Bragg, Jeff Hammerbacher, Doug Downey, Joseph~Chee Chang, and David Sontag. 2023.
\newblock \href {https://arxiv.org/abs/2304.02623} {Beyond summarization: Designing ai support for real-world expository writing tasks}.
\newblock \emph{Preprint}, arXiv:2304.02623.

\bibitem[{Skarlinski et~al.(2024)Skarlinski, Cox, Laurent, Braza, Hinks, Hammerling, Ponnapati, Rodriques, and White}]{skarlinski2024languageagentsachievesuperhuman}
Michael~D. Skarlinski, Sam Cox, Jon~M. Laurent, James~D. Braza, Michaela Hinks, Michael~J. Hammerling, Manvitha Ponnapati, Samuel~G. Rodriques, and Andrew~D. White. 2024.
\newblock \href {https://arxiv.org/abs/2409.13740} {Language agents achieve superhuman synthesis of scientific knowledge}.
\newblock \emph{Preprint}, arXiv:2409.13740.

\bibitem[{Spink et~al.(1998)Spink, Greisdorf, and Bateman}]{spink1998highly}
Amanda Spink, Howard Greisdorf, and Judy Bateman. 1998.
\newblock From highly relevant to not relevant: examining different regions of relevance.
\newblock \emph{Information processing \& management}, 34(5):599--621.

\bibitem[{Su et~al.(2022)Su, Li, Zhang, Shang, Jiang, Liu, and Fung}]{su2022readgeneratefaithfullong}
Dan Su, Xiaoguang Li, Jindi Zhang, Lifeng Shang, Xin Jiang, Qun Liu, and Pascale Fung. 2022.
\newblock \href {https://arxiv.org/abs/2203.00343} {Read before generate! faithful long form question answering with machine reading}.
\newblock \emph{Preprint}, arXiv:2203.00343.

\bibitem[{Tan et~al.(2024)Tan, Guo, Shi, Xu, Liu, Feng, Li, Wang, Shang, Liu, and Song}]{tan2024proxyqaalternativeframeworkevaluating}
Haochen Tan, Zhijiang Guo, Zhan Shi, Lu~Xu, Zhili Liu, Yunlong Feng, Xiaoguang Li, Yasheng Wang, Lifeng Shang, Qun Liu, and Linqi Song. 2024.
\newblock \href {https://arxiv.org/abs/2401.15042} {Proxyqa: An alternative framework for evaluating long-form text generation with large language models}.
\newblock \emph{Preprint}, arXiv:2401.15042.

\bibitem[{Team(2024)}]{qwen2.5}
Qwen Team. 2024.
\newblock \href {https://qwenlm.github.io/blog/qwen2.5/} {Qwen2.5: A party of foundation models}.

\bibitem[{Ujwal et~al.(2024)Ujwal, Surampudi, Mitra, and Saha}]{ujwal2024reasoning}
Utkarsh Ujwal, Sai Sri~Harsha Surampudi, Sayantan Mitra, and Tulika Saha. 2024.
\newblock " reasoning before responding": Towards legal long-form question answering with interpretability.
\newblock In \emph{Proceedings of the 33rd ACM International Conference on Information and Knowledge Management}, pages 4922--4930.

\bibitem[{Wang et~al.(2024{\natexlab{a}})Wang, Ni, Liu, Lu, Chen, Feng, Wei, Qu, Alinejad-Rokny, Lin, and Yang}]{wang2024autopatentmultiagentframeworkautomatic}
Qiyao Wang, Shiwen Ni, Huaren Liu, Shule Lu, Guhong Chen, Xi~Feng, Chi Wei, Qiang Qu, Hamid Alinejad-Rokny, Yuan Lin, and Min Yang. 2024{\natexlab{a}}.
\newblock \href {https://arxiv.org/abs/2412.09796} {Autopatent: A multi-agent framework for automatic patent generation}.
\newblock \emph{Preprint}, arXiv:2412.09796.

\bibitem[{Wang et~al.(2024{\natexlab{b}})Wang, Chen, Jia, Wang, Fang, Wang, Gao, Xie, Xu, Dai et~al.}]{wang2024weaver}
Tiannan Wang, Jiamin Chen, Qingrui Jia, Shuai Wang, Ruoyu Fang, Huilin Wang, Zhaowei Gao, Chunzhao Xie, Chuou Xu, Jihong Dai, et~al. 2024{\natexlab{b}}.
\newblock Weaver: Foundation models for creative writing.
\newblock \emph{arXiv preprint arXiv:2401.17268}.

\bibitem[{Wang et~al.(2024{\natexlab{c}})Wang, Guo, Yao, Zhang, Zhang, Wu, Zhang, Dai, Zhang, Wen, Ye, Zhang, and Zhang}]{2024autosurvey}
Yidong Wang, Qi~Guo, Wenjin Yao, Hongbo Zhang, Xin Zhang, Zhen Wu, Meishan Zhang, Xinyu Dai, Min Zhang, Qingsong Wen, Wei Ye, Shikun Zhang, and Yue Zhang. 2024{\natexlab{c}}.
\newblock Autosurvey: Large language models can automatically write surveys.
\newblock In \emph{The Thirty-eighth Annual Conference on Neural Information Processing Systems}.

\bibitem[{Wei et~al.(2024)Wei, Yang, Song, Lu, Hu, Tran, Peng, Liu, Huang, Du et~al.}]{wei2024long}
Jerry Wei, Chengrun Yang, Xinying Song, Yifeng Lu, Nathan Hu, Dustin Tran, Daiyi Peng, Ruibo Liu, Da~Huang, Cosmo Du, et~al. 2024.
\newblock Long-form factuality in large language models.
\newblock \emph{arXiv preprint arXiv:2403.18802}.

\bibitem[{Wu et~al.(2025{\natexlab{a}})Wu, Yin, Jiang, Wang, Xi, Fang, Zhou, Xie, and Huang}]{wu2025webwalker}
Jialong Wu, Wenbiao Yin, Yong Jiang, Zhenglin Wang, Zekun Xi, Runnan Fang, Deyu Zhou, Pengjun Xie, and Fei Huang. 2025{\natexlab{a}}.
\newblock \href {https://arxiv.org/abs/2501.07572} {Webwalker: Benchmarking llms in web traversal}.
\newblock \emph{Preprint}, arXiv:2501.07572.

\bibitem[{Wu et~al.(2025{\natexlab{b}})Wu, Zhu, and Liu}]{wu2025agenticreasoningreasoningllms}
Junde Wu, Jiayuan Zhu, and Yuyuan Liu. 2025{\natexlab{b}}.
\newblock \href {https://arxiv.org/abs/2502.04644} {Agentic reasoning: Reasoning llms with tools for the deep research}.
\newblock \emph{Preprint}, arXiv:2502.04644.

\bibitem[{Xia et~al.(2024)Xia, Zhu, Li, Zhu, Li, Li, Zhang, and Yao}]{xia-etal-2024-rule}
Peng Xia, Kangyu Zhu, Haoran Li, Hongtu Zhu, Yun Li, Gang Li, Linjun Zhang, and Huaxiu Yao. 2024.
\newblock \href {https://doi.org/10.18653/v1/2024.emnlp-main.62} {{RULE}: Reliable multimodal {RAG} for factuality in medical vision language models}.
\newblock In \emph{Proceedings of the 2024 Conference on Empirical Methods in Natural Language Processing}, pages 1081--1093, Miami, Florida, USA. Association for Computational Linguistics.

\bibitem[{Xu and Reitter(2017)}]{said}
Yang Xu and David Reitter. 2017.
\newblock \href {https://doi.org/10.18653/v1/P17-1058} {Spectral analysis of information density in dialogue predicts collaborative task performance}.
\newblock In \emph{Proceedings of the 55th Annual Meeting of the Association for Computational Linguistics (Volume 1: Long Papers)}, pages 623--633, Vancouver, Canada. Association for Computational Linguistics.

\bibitem[{Yang et~al.(2023)Yang, Klein, Peng, and Tian}]{yang2023docimprovinglongstory}
Kevin Yang, Dan Klein, Nanyun Peng, and Yuandong Tian. 2023.
\newblock \href {https://arxiv.org/abs/2212.10077} {Doc: Improving long story coherence with detailed outline control}.
\newblock \emph{Preprint}, arXiv:2212.10077.

\bibitem[{Zhang et~al.(2019)Zhang, Guo, Fan, Lan, and Cheng}]{zhang2019outline}
Ruqing Zhang, Jiafeng Guo, Yixing Fan, Yanyan Lan, and Xueqi Cheng. 2019.
\newblock Outline generation: Understanding the inherent content structure of documents.
\newblock In \emph{Proceedings of the 42nd International ACM SIGIR Conference on Research and Development in Information Retrieval}, pages 745--754.

\bibitem[{Zhao et~al.(2024)Zhao, Zhou, Li, Tang, Wang, Hou, Min, Zhang, Zhang, Dong, Du, Yang, Chen, Chen, Jiang, Ren, Li, Tang, Liu, Liu, Nie, and Wen}]{zhao2024surveylargelanguagemodels}
Wayne~Xin Zhao, Kun Zhou, Junyi Li, Tianyi Tang, Xiaolei Wang, Yupeng Hou, Yingqian Min, Beichen Zhang, Junjie Zhang, Zican Dong, Yifan Du, Chen Yang, Yushuo Chen, Zhipeng Chen, Jinhao Jiang, Ruiyang Ren, Yifan Li, Xinyu Tang, Zikang Liu, Peiyu Liu, Jian-Yun Nie, and Ji-Rong Wen. 2024.
\newblock \href {https://arxiv.org/abs/2303.18223} {A survey of large language models}.
\newblock \emph{Preprint}, arXiv:2303.18223.

\bibitem[{Zheng et~al.(2023)Zheng, Sharan, Jaiswal, Wang, Xi, Xu, and Wang}]{zheng2023outline}
Wenqing Zheng, SP~Sharan, Ajay~Kumar Jaiswal, Kevin Wang, Yihan Xi, Dejia Xu, and Zhangyang Wang. 2023.
\newblock Outline, then details: Syntactically guided coarse-to-fine code generation.
\newblock In \emph{International Conference on Machine Learning}, pages 42403--42419. PMLR.

\bibitem[{Zhou et~al.(2023)Zhou, Jiang, Cui, Wang, Xiao, Hou, Cotterell, and Sachan}]{zhou2023recurrentgptinteractivegenerationarbitrarily}
Wangchunshu Zhou, Yuchen~Eleanor Jiang, Peng Cui, Tiannan Wang, Zhenxin Xiao, Yifan Hou, Ryan Cotterell, and Mrinmaya Sachan. 2023.
\newblock \href {https://arxiv.org/abs/2305.13304} {Recurrentgpt: Interactive generation of (arbitrarily) long text}.
\newblock \emph{Preprint}, arXiv:2305.13304.

\end{thebibliography}

\appendix
\clearpage

\section{OmniThink Details}
\subsection{Implementation}
\label{app:implementation}
We build OmniThink based on the DSpy framework~\cite{khattab2023dspycompilingdeclarativelanguage}, and STORM.
Appendix~\ref{appendix:fullprompt} contains the corresponding prompts we used. 
During article generation, we set the \textit{temperature} at 1.0 and \textit{top\_p} at 0.9. 
The search engine employed is Bing's API, with the parameter for the number of web pages returned per query configured to 5.
To retrieve information based on the outline, we use SentenceBERT~\cite{reimers-gurevych-2019-sentence} embeddings to calculate cosine similarity, thereby retrieving the three most similar web pages each time.
For the computation of knowledge density, we utilize Factscore\footnote{\url{https://github.com/shmsw25/FActScore}} with GPT-4o-08-06 as the backbone to decompose atomic knowledge \cite{min2023factscorefinegrainedatomicevaluation}. 
After the decomposition, we proceed to use GPT-4o-08-06 for the deduplication of the split atomic knowledge.

\subsection{Full Prompts in OmniThink}
\label{appendix:fullprompt}

In \S \ref{sec:omnithink}, we introduce the specific process of OmniThink, which is implemented using zero-shot prompting based on GPT-4o-2024-08-06. Lists \ref{lst:expand_prompt}, \ref{lst:reflect_prompt}, \ref{lst:outline_prompt}, \ref{lst:section_prompt} and \ref{lst:polish_prompt}, respectively document the complete prompts for OmniThink's Expand, Reflect, Write Outline, Write Article, and Polish Article stages.
These prompts are designed to guide the model through iterative stages of content generation, ensuring coherence and depth in the produced text. 

The structured process leverages dynamic adjustments based on intermediate outputs, reflecting a balanced integration of retrieval and generation capabilities. 
This systematic approach highlights OmniThink's ability to adaptively construct well-organized and contextually relevant articles across diverse topics.

\begin{figure}[]
    \centering
    \includegraphics[width=0.85\linewidth]{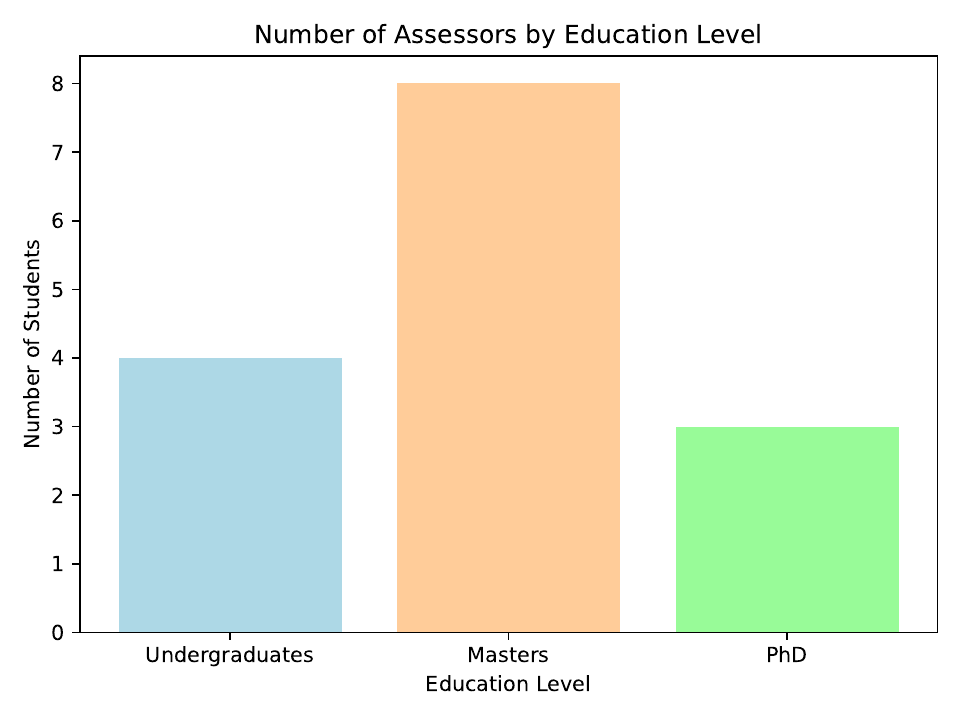}
    \caption{The educational background distribution of assessors.
}
    \label{fig:education}
\end{figure}

\section{Automatic Evaluation Details}
\label{appendix:autoeval}
To further ensure reliability, we conducted multiple evaluation rounds using different prompts covering various aspects of outline coherence, structural logic, and topic relevance. 
This multi-faceted evaluation helps mitigate potential biases and enhances the robustness of the scoring results.
\subsection{Outline Evaluation}
\label{appendix:autoeval_outline}
Since Prometheus2~\cite{kim2024prometheus2opensource} does not perform targeted optimization on the outline, we decided to use a more powerful model to score the outline. 
To ensure the results are consistent, we set the temperature to 0.
Specifically, we use the Prometheus2 framework but replace the underlying evaluation model with GPT-4o-08-06. 
The scoring criteria for outline quality evaluation and discourse quality evaluation can be found in Lstlisting \ref{lst:outlinerubric}.
In addition, since Co-STORM does not have an intermediate outline generation step, we had to extract the outline from the final article for evaluation, which might be the reason for the relatively lower outline scores observed form Co-STORM.

\subsection{Article Evaluation}
\label{appendix:autoeval_article}
Following Co-STORM~\cite{jiang2024unknownunknowns}, we utilized the Prometheus-7b-v2.0 model for evaluation. 
Prometheus~\cite{kim2024prometheus2opensource} is an open-source scoring model used to assess lengthy texts based on user-defined criteria. 
Its default temperature value is 1.0, and the top\_p value is 0.9. 
Due to the model's limited context window, we exclude reference sections from the article evaluation and trim the input text to fewer than 2000 words to fit within the model's context window. 
This is consistent with STORM's approach~\cite{shao2024assisting}, where the shortest section is removed each time until the article length meets the specified requirement. 
The scoring criteria for article quality evaluation can be found in Listing \ref{lst:rubric}.
\subsection{Variance of Article Evaluation}
\label{appendix:variance}
As shown in the table \ref{table:variance}, we present the variance of three evaluation runs using the previously saved checkpoints on Prometheus-7B-v2.0.Thanks to the solid alignment of Prometheus-7B-v2.0, the variances are relatively small.
\begin{table}[h]
\centering
\resizebox{\columnwidth}{!}{
\begin{tabular}{lcccc}
\hline
\textbf{Method} & \textbf{Relevance} & \textbf{Breadth} & \textbf{Depth} & \textbf{Novelty} \\
\hline
RAG        & 0.0027 & 0.0060 & 0.0092 & 0.0073 \\
oRAG       & 0.0043 & 0.0071 & 0.0111 & 0.0132 \\
STORM      & 0.0027 & 0.0052 & 0.0021 & 0.0085 \\
Co-STORM   & 0.0032 & 0.0066 & 0.0036 & 0.0106 \\
OmniThink  & 0.0011 & 0.0027 & 0.0042 & 0.0095 \\
\hline
\end{tabular}
}
\caption{Variance of three evaluation on Prometheus-7B-v2.0}
\label{table:variance}
\end{table}

\section{Human Evaluation}
\label{appendix:huamneval}

\subsection{Human Evaluation Details}
We randomly select 20 topics and compare articles generated by our method with those from the Co-STORM (the comprehensive best-performing baseline based on automatic evaluation), scoring them on the same four aspects. 
The participants in the evaluation voluntarily provided their highest educational qualification to demonstrate their ability to impartially assess the article.
As shown in Figure \ref{fig:education}, all of our human evaluators have an undergraduate degree or higher, with 53\% having a graduate degree.
As discussed in \S \ref{appendix:huamneval}, to compare the merits of OmniThink and Co-STORM, each human evaluator was given a scoring criterion and a pair of articles. 
They were required to compare and assign scores, with the scoring criteria being the same as Lstlisting \ref{lst:rubric}. 
We compiled the average scores given by the human evaluators for OmniThink and Co-STORM and compared their wins and losses.

\subsection{Human Evaluation Results}
\begin{figure}[h]
    \centering
    \includegraphics[width=\linewidth]{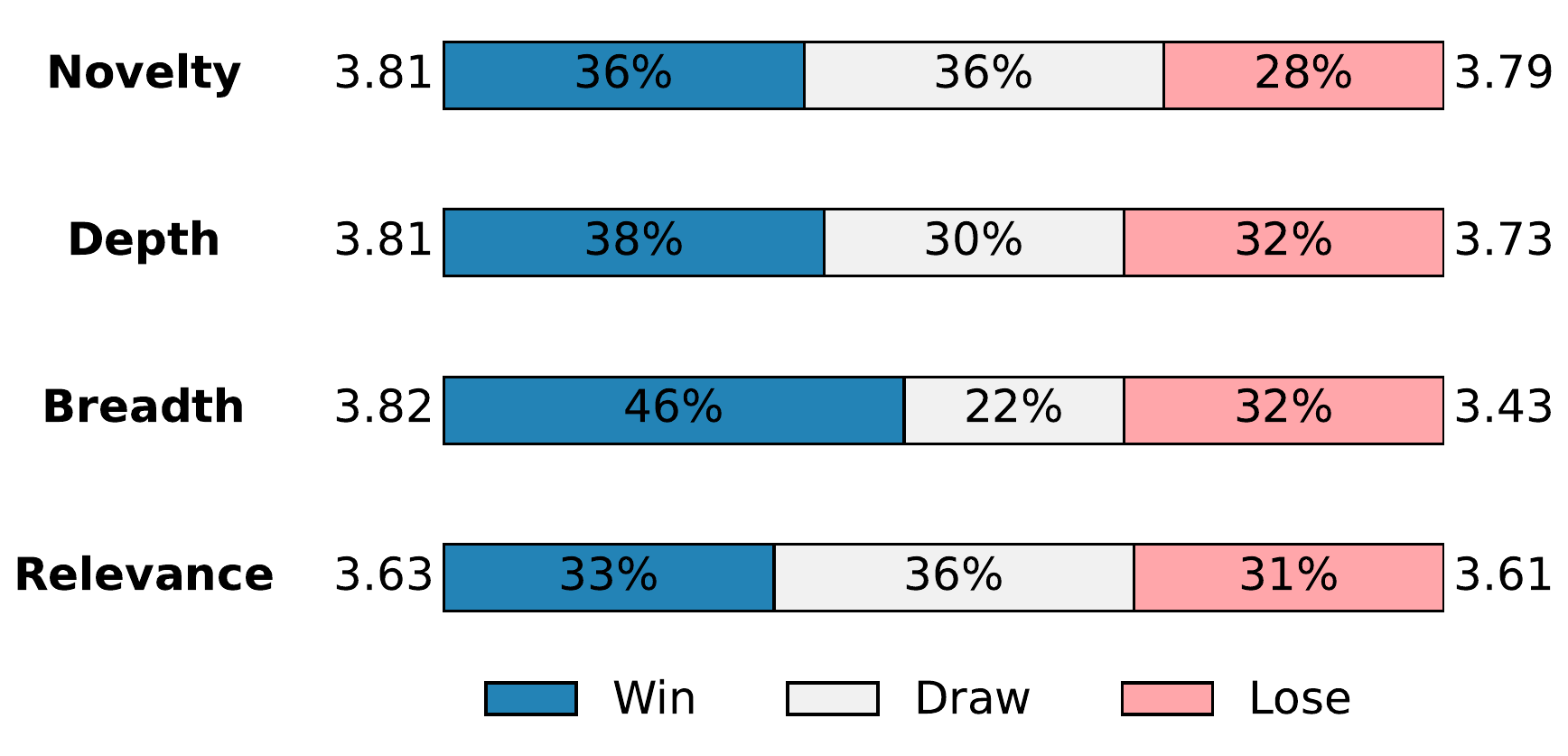}
    \caption{
    Comparison of OmniThink and Co-STORM results under human evaluation.
    The values on the left side represent the average score from OmniThink human evaluators, while the values on the right side represent the average score from Co-STORM human evaluators.
}
    \label{fig:human}
\end{figure}
To better understand the strengths and weaknesses of OmniThink, we engage 15 well-educated volunteers to conduct a human evaluation. 
In Figure \ref{fig:human}, we present the results of human scoring. 
The findings indicate that OmniThink's average performance surpasses that of the current strongest baseline across various dimensions, with a notable 11\% improvement in the Breadth metric compared to Co-STORM. 
However, in terms of the Novelty metric, although automated evaluation shows an 11\% enhancement, human assessment reveals only a marginal advantage. 
This discrepancy suggests that the current automated evaluation may not yet be fully aligned with human judgment, highlighting a direction for future improvement in the evaluation of long texts.

It should also be noted that despite OmniThink's overall superior performance in various dimensions, approximately 30\% of the articles are considered equally excellent to the baseline by human evaluators. 
This could be attributed to the increasing difficulty for humans to discern subtle differences as the foundational writing capabilities of large models improve.
Consequently, there is an urgent need to develop more rigorous and fine-grained evaluation methods to assess model performance more accurately.

\section{Further Analysis}
\subsection{Unique URL Analysis}
To further investigate whether OmniThink surpasses these predefined boundaries, we conduct an unique url experiment. 
The goal is to examine whether OmniThink can retrieve more unique URLs compared to other methods, thus enabling the generation of more diverse and innovative content.
\begin{table}[h!]
\centering
\resizebox{\columnwidth}{!}{
    \begin{tabular}{lcccc}
    \hline
    Method       & OmniThink & Co-STORM & STORM & oRAG  \\ \hline
    Unique URLs  & 120.63    & 10.49    & 16.56 & 2.15  \\ \hline
    \end{tabular}
}
\caption{Average number of unique URLs retrieved by each method.}
\label{table:url}
\end{table}

Table \ref{table:url} show that OmniThink retrieves significantly more unique URLs compared to other methods, such as Co-STORM, STORM, and oRAG. This indicates that OmniThink can access a broader range of diverse web content, which in turn enables the generation of more innovative and in-depth articles.

\subsection{Processing Time Analysis}
We have recorded the time required for each method to run in the main table. Based on cost considerations, we use Google Search and Qwen-Plus. 
We ran 10 cases for each and calculated the average time taken.
\begin{table}[h!]
\centering
\resizebox{\columnwidth}{!}{
    \begin{tabular}{lccc}
    \hline
    Method       & OmniThink & Co-STORM & STORM  \\ \hline
    time(s)  & 322    & 289   & 289  \\ \hline
    \end{tabular}
}
\caption{Average time taken by each method.}
\label{table:time}
\end{table}

As shown in Table \ref{table:time}, the current state of long text generation has encountered a certain bottleneck.
We bypassed the scaling of complex text writing pipelines and instead focused on scaling from the data perspective to enhance text quality.
We embraced the current trend of multiple rounds of reflection, led by DeepResearch. Therefore, we believe that these processing time costs are worthwhile.

\section{Information Boundary Experiments Details}
\label{appendix:informationboundary}
In the information boundary analysis, our data comes from the results in Table 1, based on GPT-4o as the backbone.
we extract the snippets content of each retrieved webpage from the search engine, then use Sentence-BERT to extract their representations.
After reducing the dimensions to a 2D plane using PCA, we apply normalization and calculate the centroid for each category. 
Outliers, defined as points beyond 1.5 times the standard deviation, are exclude, and the convex hull formed by the remaining points is computed.

\section{Expansion \& Reflection Experiments Details}
\label{appendix:er}
Given the interdependent nature of expansion and reflection in OmniThink, it is impractical to assess their individual impacts in isolation. 
To address this challenge, we adopt an indirect yet systematic approach to evaluate their collective influence on the final articles' quality.
During the information acquisition phase, we substitute the model used for expansion with a lower-performing model and measured the extent of performance decline in the generated article's metrics, which served as an indicator of the impact of the expansion process on these metrics.
Specifically, based on the experimental results for qwen-plus-2024-08-06, we replace the models used for the expansion and reflection processes from Qwen-Plus to Qwen2.5-7b-instruct~\cite{qwen2.5} and observe the decline in various evaluation results. 
This transition allows us to observe and document the subsequent changes in a range of evaluation metrics, providing insights into the expansion and reflection process's influence on the articles' overall assessment.

\section{Comparison of features across different methods}
\label{appendix:compare}
\paragraph{Dynamic retrieval}
In previous methods, STORM and Co-STORM primarily retrieve web pages through ongoing dialogue, largely relying on the maximum number of conversations, without dynamically adjusting the retrieval of web content according to the difficulty and depth of the problem. 
OmniThink achieves dynamic retrieval based on the problem's difficulty by constantly reflecting on whether further retrieval is necessary with the current content.
\paragraph{Structured memory}
STORM stores web content merely through dialogue, while Co-STORM records a mind map during the conversation process. 
OmniThink not only records retrieved web pages in a progressive knowledge manner but also uses a conceptual pool to document changes in the LLM's understanding of the topic.
\paragraph{Reflective thinking}
In STORM and Co-STORM, continuous dialogue mainly occurs through role-playing, without reflection on the retrieved content. 
OmniThink achieves better results by continuously reflecting on the retrieved content to fill the conceptual pool.

\section{Effectiveness of Knowledge Density}
\label{appendix:kd}

We designed an interesting experiment to demonstrate the effectiveness of the KD evaluation metric.

First, we constructed 50 unique atomic facts across different topics and asked GPT-4o to generate a ~500-word article based on these facts.
Then, we gradually reduced the number of atomic facts while keeping the article length unchanged, in order to simulate articles with varying levels of knowledge density.
To ensure stylistic consistency, all generations were produced using GPT-4o, so that the articles remained largely consistent in expression apart from differences in knowledge density.
We invited three human volunteers and three language model evaluators (GPT-4o, DeepSeek-R1, and O3-mini-high) to assign preference scores to the articles generated with different amounts of atomic knowledge.
The experimental results are shown in Table \ref{tab:kd}.
\begin{table}[h]
\centering
\resizebox{\linewidth}{!}{%
\begin{tabular}{lcccc}
\toprule
\textbf{Evaluator} & \textbf{20 Facts} & \textbf{30 Facts} & \textbf{40 Facts} & \textbf{50 Facts} \\
\midrule
GPT-4o          & 1.0 & 2.0 & 3.0 & 3.6 \\
DeepSeek-R1     & 1.0 & 2.6 & 2.8 & 3.6 \\
O3-mini-high    & 1.0 & 2.4 & 3.2 & 3.4 \\
Humans          & 1.0 & 2.1 & 3.1 & 3.8 \\
\bottomrule
\end{tabular}%
}
\caption{Preference scores assigned by human and LLM evaluators for articles generated with varying numbers of atomic facts.}
\label{tab:kd}
\end{table}

\section{Case Study}
In Figure \ref{fig:agi40}, we present an example of AGI generated by OmniThink.
It is generated using GPT-4o as the backbone. 
We can see that OmniThink's language is more concise compared to other methods, and it contains more information per unit of text length.

In addition, we present an example of AGI generated by the Reasoning model in Figure \ref{fig:agir1}. We can observe that the OmniThink using the Reasoning model cites significantly more content per chapter, indicating that the model has improved its ability to utilize information through reflection.

\section{Decision Process of Expansion}
\label{appendix:decision}
\begin{algorithm}[H]
\caption{Decision Process of Expansion}
\label{alg:decision_expand}
\begin{algorithmic}[1]
    \STATE \textbf{Input:} Tree $\mathcal{T}$, Max Depth $D$, Conceptual Pool $\mathcal{P}$
    \STATE \textbf{Output:} Updated $\mathcal{T}$ and $\mathcal{P}$

    \WHILE{$\text{depth}(\mathcal{T}) < D$}
        \FOR{each leaf node $N_i$ in $\mathcal{T}$}
            \STATE $R_i \gets \text{LLM.decide\_next}(\mathcal{P}, N_i)$
            \IF{$R_i$ requires expansion}
                \STATE Extract keywords and retrieve info
                \STATE Create sub-nodes and add to $\mathcal{T}$
            \ENDIF
        \ENDFOR
        \STATE Update $\mathcal{P}$ with new insights
        \IF{early stopping condition met}
            \STATE \textbf{break}
        \ENDIF
    \ENDWHILE

    \STATE \textbf{Return} $\mathcal{T}, \mathcal{P}$
\end{algorithmic}
\end{algorithm}

In practice, we first check whether each leaf node of the information tree has reached a predefined maximum depth. If it has not, we feed the content and type of that node, along with the current conceptual pool, to the LLM as a prompt. The LLM is instructed to decide whether the node requires further expansion. If expansion is needed, the model generates potential sub-node categories and corresponding retrieval keywords based on the conceptual pool; otherwise, if the node is deemed sufficiently complete, the model produces no output.

To operationalize this, we extract the sub-node categories and keywords from the model's response using regular expressions. 
These elements are then employed to query web search engines or retrieval systems. 
The retrieved content forms the basis of new information nodes, which are added to the current information tree to iteratively refine and expand the knowledge structure.

Algorithm \ref{alg:decision_expand} is a brief pseudocode illustrating the overall expansion process.
We first check whether the information tree has reached a predefined maximum depth. 
If not, the LLM is queried to decide the next steps for each leaf node. 
New information is retrieved accordingly and integrated into the tree. The conceptual pool is also dynamically updated during the expansion process.

\section{Clarification of Reflection}
In this paper, our reflection refers to the process where the LLM reflects on the retrieved information based on its current Conceptual Pool, evaluating which parts of the information can enrich the existing Conceptual Pool. 
The usable information is then extracted as insights and added to the Conceptual Pool.

\section{Pseudo-code of Expansion \& Reflection}

\begin{algorithm}[H]
\caption{Expansion and Reflection}
\label{alg:OmniThink}
\begin{algorithmic}[1]
    \STATE \textbf{Input:} Topic $\mathrm{T}$, Depth $K$
    \STATE \textbf{Output:} Information Tree $\mathcal{T}$, Conceptual Pool $\mathcal{P}$

    \COMMENT{\textbf{Initialization}}
    \STATE Initialize Information Tree $\mathcal{T}_0$ with root node $N_r$
    \STATE Retrieve initial information using search engines
    \STATE Organize and analyze information to form Conceptual Pool $\mathcal{P}_0$
    
    \COMMENT{\textbf{Expansion and Reflection}}
    \FOR{each time step $m = 0$ to $K - 1$}
        \STATE $L_m \gets$ Leaf Nodes of $\mathcal{T}_m$
        \STATE Store $L_m$ in Conceptual Buffer $\mathcal{P}_b$
        \FOR{each node $N_i$ in $L_m$}
            \IF{Needs Expansion($N_i$)}
                \STATE Determine expansion areas using $\mathcal{P}_m$
                \STATE Generate sub-nodes $\text{SUB}(N_i) = \{ S_0, S_1, \ldots, S_{k_{N_i}} \}$
                \FOR{each sub-node $S_j$ in $\text{SUB}(N_i)$}
                    \STATE Retrieve information for $S_j$
                    \STATE Add $S_j$ to $\mathcal{T}_{m+1}$
                \ENDFOR
            \ENDIF
        \ENDFOR

        \STATE $L_{m+1} \gets$ Leaf Nodes of $\mathcal{T}_{m+1}$
        \STATE Analyze, filter, and synthesize information from $L_{m+1}$ to obtain insights $I_{m+1}$
        \STATE Update Conceptual Pool $\mathcal{P}_{m+1} \gets \text{Merge}(I_{m+1}, \mathcal{P}_m)$

        \IF{Sufficient information acquired}
            \STATE \textbf{break}
        \ENDIF
    \ENDFOR

    \STATE \textbf{Return} Final Article $\mathcal{A}$
\end{algorithmic}
\end{algorithm}

\begin{figure*}
\input{prompt/expand}
\captionof{lstlisting}{Prompts used for expanding in OmniThink.}
\label{lst:expand_prompt}
\end{figure*}

\begin{figure*}
\input{prompt/reflect}
\captionof{lstlisting}{Prompts used for reflecting in OmniThink.}
\label{lst:reflect_prompt}
\end{figure*}

\begin{figure*}
\input{prompt/outline}
\captionof{lstlisting}{Prompts used for writing the outline in OmniThink.}
\label{lst:outline_prompt}
\end{figure*}

\begin{figure*}
\input{prompt/writesection}
\captionof{lstlisting}{Prompts used for writing section in OmniThink.}
\label{lst:section_prompt}
\end{figure*}

\begin{figure*}
\input{prompt/polish}
\captionof{lstlisting}{Prompts used for polishing article in OmniThink.}
\label{lst:polish_prompt}
\end{figure*}

\begin{figure*}
\resizebox{\textwidth}{!}{
    \begin{tabular}{ll} 
\toprule
Criteria Description & \textbf{Guidance for Content Generation}: \makecell[l]{Does the outline effectively guide content generation, ensuring comprehensive coverage of the topic?} \\
Score 1 Description & The outline fails to guide content generation, omitting significant aspects of the topic or providing insufficient direction. \\
Score 2 Description & The outline provides limited guidance, covering some key areas but lacking depth or completeness in addressing the topic. \\
Score 3 Description & The outline provides moderate guidance for content generation, addressing most key areas but leaving some gaps or ambiguities. \\
Score 4 Description & The outline effectively guides content generation, covering all significant aspects with clear direction, though minor refinements could enhance comprehensiveness. \\
Score 5 Description & The outline is exemplary in guiding content generation, thoroughly addressing all aspects of the topic with clear, detailed direction and no significant gaps. \\
\midrule
Criteria Description & \textbf{Hierarchical Clarity}: \makecell[l]{Does the outline clearly define a hierarchy of topics and subtopics, with a logical, diverse structure that is easy to understand?} \\
Score 1 Description &  \makecell[l]{The outline exhibits no discernible hierarchical structure. \\Topics and subtopics are jumbled together without logical separation or clear levels, making it nearly impossible to follow or identify any organization.} \\
Score 2 Description & \makecell[l]{The outline attempts to establish a hierarchy but fails to maintain logical consistency. Main topics and subtopics are frequently misclassified,\\ and the structure is overly rigid or disjointed. Subtopics may be missing, misplaced, or redundant, making it hard to grasp the intent of the structure.} \\
Score 3 Description & \makecell[l]{The outline has a recognizable hierarchical structure but lacks diversity in organization style. While main topics are somewhat clear, subtopics occasionally overlap, \\are misaligned, or follow a repetitive format. This restricts flexibility and introduces mild confusion in certain areas.} \\
Score 4 Description & \makecell[l]{The outline displays a clear, logical, and diverse hierarchical structure. Main topics are distinct, and subtopics are properly nested. While most elements are well-placed, \\there may be minor redundancies or opportunities to introduce more diverse formats for subtopics. Slight adjustments could achieve better precision and variety in style.} \\
Score 5 Description & \makecell[l]{The outline showcases an exceptional, flawless hierarchical structure. Each main topic is distinct, and subtopics are logically nested with absolute clarity and stylistic diversity.\\ The outline demonstrates flexibility in structure and organization, adapting its style where appropriate for the content and logic. No further refinement is necessary.} \\
\midrule
Criteria Description & \textbf{Logical Coherence}: \makecell[l]{Does the outline logically organize topics and subtopics, ensuring a smooth and natural flow of ideas with clear logical transitions?} \\
Score 1 Description & \makecell[l]{The outline is highly disjointed and incoherent. Topics and subtopics appear in a random, unordered manner, with no logical flow or sense of progression. \\Major conceptual gaps and illogical jumps are present throughout the structure.} \\
Score 2 Description & \makecell[l]{The outline shows some attempt at logical organization, but it contains frequent inconsistencies, abrupt shifts, or logical missteps. \\Topics and subtopics are misaligned or lack proper transitions, making the reader work hard to follow the structure.} \\
Score 3 Description & \makecell[l]{The outline demonstrates a basic level of logical coherence. Most topics follow a general sequence, but some sections feel forced, with weak or unclear transitions.\\ There are small jumps in logic, causing slight confusion or loss of flow at certain points.} \\
Score 4 Description & \makecell[l]{The outline exhibits a strong sense of logical flow, with ideas presented in a mostly smooth and connected manner.\\ Transitions between topics and subtopics are clear, but a few minor adjustments could make the flow more seamless or natural. The logic is sound, but room for refinement exists.} \\
Score 5 Description & \makecell[l]{The outline achieves exceptional logical coherence. Each topic and subtopic follows a deliberate, thoughtful progression, with clear, natural, and intuitive transitions.\\ The reader experiences a seamless flow of ideas, and no adjustments are required to improve logical consistency or flow.} \\
\bottomrule
\end{tabular}

}
\caption{Outline scoring rubrics on a 1-5 scale for the Prometheus model.}
\label{lst:outlinerubric}
\end{figure*}

\begin{figure*}
\resizebox{\textwidth}{!}{
    \begin{tabular}{ll} 
\toprule
Criteria Description & \textbf{Broad Coverage}: Does the article provide an in-depth exploration of the topic and have good coverage?                                                                                                               \\
Score 1 Description  & Severely lacking; offers little to no coverage of the topic's primary aspects, resulting in a very narrow perspective.                                                                                                                        \\
Score 2 Description  & Partial coverage; includes some of the topic's main aspects but misses others, resulting in an incomplete portrayal.                                                                                                                              \\
Score 3 Description  & Acceptable breadth; covers most main aspects, though it may stray into minor unnecessary details or overlook some relevant points.                                                                                             \\
Score 4 Description  & Good coverage; achieves broad coverage of the topic, hitting on all major points with minimal extraneous information.                                                                      \\
Score 5 Description  & Exemplary in breadth; delivers outstanding coverage, thoroughly detailing all crucial aspects of the topic without including irrelevant information.                                                                                \\ 
\midrule
Criteria Description & \textbf{Novelty}: Does the report cover novel aspects that relate to the user's initial intent but are not directly derived from it?                                                                                              \\
Score 1 Description  & Lacks novelty; the report strictly follows the user's initial intent with no additional insights.                                                                                                                                     \\
Score 2 Description  & Minimal novelty; includes few new aspects but they are not significantly related to the initial intent.                                                                                              \\
Score 3 Description  & Moderate novelty; introduces some new aspects that are somewhat related to the initial intent.                                                                                  \\
Score 4 Description  & Good novelty; covers several new aspects that enhance the understanding of the initial intent.                                                                                                                      \\
Score 5 Description  & Excellent novelty; introduces numerous new aspects that are highly relevant and significantly enrich the initial intent.                                                                             \\ 
\midrule
Criteria Description & \textbf{Relevance and Focus}: How effectively does the report maintain relevance and focus, given the dynamic nature of the discourse?                                                                  \\
Score 1 Description  & Very poor focus; discourse diverges significantly from the initial topic and intent with many irrelevant detours.                                                                                                \\
Score 2 Description  & Poor focus; some relevant information, but many sections diverge from the initial topic.                                                                     \\
Score 3 Description  & Moderate focus; mostly stays on topic with occasional digressions that still provide useful information.                                                                                                                                    \\
Score 4 Description  & Good focus; maintains relevance and focus throughout the discourse with minor divergences that add value.                                                                 \\
Score 5 Description  & Excellent focus; consistently relevant and focused discourse, even when exploring divergent but highly pertinent aspects.  \\ 
\midrule
Criteria Description & \textbf{Depth of Exploration}: How thoroughly does the report explore the initial topic and its related areas, reflecting the dynamic discourse?                                                                     \\
Score 1 Description  & Very superficial; provides only a basic overview with significant gaps in exploration.                                                                    \\
Score 2 Description  & Superficial; offers some detail but leaves many important aspects unexplored.                                                                    \\
Score 3 Description  & Moderate depth; covers key aspects but may lack detailed exploration in some areas.                                                   \\
Score 4 Description  & Good depth; explores most aspects in detail with minor gaps.                                                                   \\
Score 5 Description  & Excellent depth; thoroughly explores all relevant aspects with comprehensive detail, reflecting a deep and dynamic discourse.                               \\
\bottomrule
\end{tabular}
}
\caption{Report scoring rubrics on a 1-5 scale for the Prometheus model.}
\label{lst:rubric}
\end{figure*}

\begin{figure*}[h]
    \centering
    \input{case/agi4o}
    \caption{A case of AGI generated by OmniThink with GPT-4o.
}
    \label{fig:agi40}
\end{figure*}

\begin{figure*}[h]
    \centering
    \input{case/agir1}
    \caption{A case of AGI generated by OmniThink with DeepSeek-R1.
}
    \label{fig:agir1}
\end{figure*}

\end{document}